\documentclass[10pt,twocolumn,letterpaper]{article}

\usepackage[pagenumbers]{cvpr} %

\usepackage{tikz}

\usepackage{colortbl}
\usepackage{multirow}
\definecolor{rankfirst}{rgb}{1,0.70,0.70}
\definecolor{ranksecond}{rgb}{1,0.85,0.70}
\definecolor{rankthird}{rgb}{1,1,0.70}
\newcommand{\fst}{\cellcolor{rankfirst}}
\newcommand{\snd}{\cellcolor{ranksecond}}
\newcommand{\trd}{\cellcolor{rankthird}}
\DeclareRobustCommand{\rankbox}[2]{{\setlength{\fboxsep}{1pt}\colorbox{#1}{#2}}}
\DeclareRobustCommand{\rankkey}{\rankbox{rankfirst}{best}, \rankbox{ranksecond}{second} and \rankbox{rankthird}{third}}

\renewcommand{\arraystretch}{0}

\newlength{\gapsz}
\newlength{\labelw}
\newlength{\imgw}
\newlength{\psnrpad}
\newlength{\cropgap}
\newlength{\cropw}

\definecolor{cvprblue}{rgb}{0.21,0.49,0.74}
\usepackage[pagebackref,breaklinks,colorlinks,allcolors=cvprblue]{hyperref}

\def\confName{CVPR}
\def\confYear{2027}

\definecolor{costgray}{HTML}{6D7178}\definecolor{costorange}{HTML}{C95F17}\definecolor{costblue}{HTML}{2E4D73}  %
\providecommand{\refrow}[1]{\textcolor{gray}{#1}} %
\AtBeginDocument{%
\crefname{appendix}{Sec.}{Secs.}
\Crefname{appendix}{Section}{Sections}
\crefname{subappendix}{Sec.}{Secs.}
\Crefname{subappendix}{Section}{Sections}
}
\providecommand{\tbd}{\textcolor{red}{?}}

\title{Tokens as Locality-Preserving Anchors --- Scalable Novel View Synthesis with Target-Adaptive Token Routing}

\author{Sagnik Ghosh \qquad Aswin C. Sankaranarayanan\\
Carnegie Mellon University}

\begin{document}
\twocolumn[{%
\renewcommand\twocolumn[1][]{#1}%
\maketitle
\begin{center}
\centering
\includegraphics[width=\textwidth]{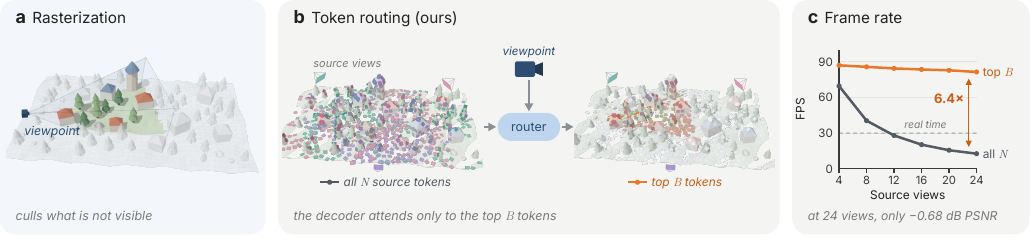}
\captionof{figure}{\textbf{Target-adaptive token routing.}
(a)~A rasterizer discards the primitives that a viewpoint does not see, so that its cost follows what is visible rather than the size of the scene.
(b)~Source tokens have no explicit position that could be tested in this way, but each of them encodes a local part of the scene, seen in one or more source views (colors).
Our router predicts from the source tokens and the viewpoint which $B$ of the $N$ source tokens the decoder needs, and the decoder attends only to these (illustration).
Selecting the tokens explicitly, with a geometric cost computed from predicted depth maps, trails our router by 1.62\,dB with 24 source views and one view's worth of tokens (\cref{sec:exp_selection}).
(c)~Frame rate against the number of source views for LagerNVS~\cite{szymanowicz2026lagernvs} with all source tokens and for our model with two views' worth of tokens, at $256\times256$ on an RTX PRO 5000: with 24 source views, our model is 6.4 times as fast, at a PSNR only 0.68\,dB lower.}
\label{fig:teaser}
\end{center}
}]

\begin{abstract}
Transformer-based view synthesis models render each target viewpoint from the source images, without an explicit 3D representation, which lets them reproduce fine detail and view-dependent effects.
However, their decoder attends to every source token for every frame, so rendering slows with each source image.
Rasterization avoids this by culling primitives that the viewpoint does not see, but source tokens cannot be culled this way, as they have no explicit position or extent.
Our key insight is that source tokens are nonetheless spatial anchors: each encodes a local region of the scene, so the decoder's attention for any viewpoint concentrates on a few of them.
A lightweight router predicts these tokens from the source tokens and the viewpoint, and the decoder attends only to a fixed budget of top-ranked tokens, so its cost no longer depends on the number of source tokens.
We supervise the router with the decoder's attention, using non-maximum suppression to demote redundant tokens.
With 24 source images and two views' worth of tokens, our model renders 6.4 times faster than LagerNVS, at a PSNR only 0.68\,dB lower.
One ranking serves every budget, so the budget can change at render time.
\end{abstract}

\section{Introduction}\label{sec:intro}
Classical rendering pipelines owe much of their speed to explicit primitives, which can be culled for a viewpoint down to the small fraction that contributes to it.
Because each primitive occupies a known region of space, a renderer can discard those that lie outside the view frustum or behind other surfaces without rasterizing them~\cite{clark1976hierarchical,greene1993hierarchical,cohenor2003survey}.
As a result, the cost of rendering a viewpoint depends on what is visible from it, not on the extent of the scene or on how many images it was reconstructed from.

Transformer-based view synthesis models~\cite{sajjadi2022srt,jin2025lvsm,szymanowicz2026lagernvs} synthesize some of the most accurate novel views from a few images~\cite{jin2025lvsm,szymanowicz2026lagernvs}: an encoder turns the source images once into source tokens, one per image patch, and a decoder renders each patch of the target image by attending to them.

Two properties of this design make rendering slow.
First, every source image adds a whole set of tokens, and the decoder attends to all of them for every viewpoint, so its cost grows with the number of source images.
Second, view synthesis relies on redundancy: a part of the scene must be seen in several source images to estimate its depth and view-dependent appearance, so the same part is encoded by tokens from several images.

Our key insight is that source tokens are locality-preserving anchors in the scene, and hence amenable to optimizations analogous to those of rasterization.
Unlike explicit primitives, however, they have no position or extent to test against the view frustum or other surfaces, so they cannot be culled directly.
The decoder's attention~\cite{szymanowicz2026lagernvs} instead identifies the tokens that a viewpoint needs, concentrating on a small fraction of them.
However, this attention is only available after running the decoder on all source tokens, which is the cost we aim to avoid.
We therefore train a router that predicts, from the source tokens and the viewpoint alone, which tokens the decoder attends to, and let the decoder attend only to the top-ranked ones within a fixed budget.
The router splits its work in two: a heavy branch processes the source tokens once per scene, like the encoder, and a light branch scores them for every viewpoint, adding little to the rendering time.

Ranking by attention mass alone does not address the second problem, since the tokens that encode the most salient regions of the viewpoint in different source images all receive high mass.
A similar problem arises in edge, keypoint and object detection, where detectors respond at many nearby locations around the same structure, and non-maximum suppression resolves it by keeping only the strongest response~\cite{4767851,bodla2017softnms}.
Source tokens from different views share no image coordinates, so we measure their overlap by their cross-attention profiles.
We discount each token's attention mass by its overlap with the tokens already selected and supervise the router with the resulting ranking, so that no suppression step runs when rendering.

Our contributions are:
\begin{itemize}[leftmargin=*,topsep=2pt,itemsep=1pt]
\item a router that ranks the source tokens for every viewpoint and processes them once per scene, so that the decoder attends only to a fixed budget of tokens and its cost barely depends on the number of source images;
\item router supervision by non-maximum suppression over the source tokens' cross-attention profiles, which gives one score per token for every budget;
\item experiments showing that, with a token budget of two views' worth and 24 source images, our model renders 6.4 times as fast as LagerNVS~\cite{szymanowicz2026lagernvs} at a PSNR 0.68\,dB lower, that selecting tokens outperforms selecting whole source views in quality and stability along a camera path, and that non-maximum suppression matters most with many views.
\end{itemize}

Because a single ranking serves every budget, the budget can change from one viewpoint to the next without re-encoding the scene.

\section{Related work}\label{sec:related}
\paragraph{Feed-forward novel view synthesis.}
Feed-forward methods have predicted multiplane images~\cite{zhou2018stereo}, neural radiance fields conditioned on source-view features~\cite{yu2020pixelnerf,mvsnerf,wang2021ibrnet} and, more recently, 3D Gaussians~\cite{kerbl20233dgs} for fast rasterization~\cite{charatan2024pixelsplat,chen2024mvsplat,zhang2024gslrm,xu2025depthsplat,chen2025longlrm,ye2025noposplat,jiang2025anysplat,ye2026yonosplat,wang2025zpressor,bui2026ecosplat}, whose explicit representation is hard to predict for thin structures, reflections and unobserved regions~\cite{szymanowicz2026lagernvs}.
Transformer-based view synthesis models~\cite{sajjadi2022srt,jin2025lvsm,jiang2025rayzer,zhang2026lact,kim2026svsm,jia2026efficientlvsm,szymanowicz2026lagernvs} instead run a large decoder for every frame and are the state of the art from a few source views~\cite{szymanowicz2026lagernvs}; our router seeks a middle ground that keeps this renderer and its quality but, like a rasterizer, uses only the tokens each viewpoint needs.

\paragraph{Token reduction and sparse attention.}
Bottleneck architectures~\cite{jin2025lvsm,kim2026svsm,park2026aimsanchorintegratedmultiviewsynthesis} and compressed representations~\cite{wang2025clift} condense the scene into fewer tokens once per scene, and CLiFT~\cite{wang2025clift} also gathers tokens by ray proximity to the target.
In two works concurrent with ours, GLoD-LVSM~\cite{lin2026glodlvsm} culls condensed tokens outside the target frustum, and TTB~\cite{lin2026ttb} bakes the cross-view attention into a small network.
Token pruning and merging in vision transformers~\cite{bolya2023tome,rao2021dynamicvit,liang2022evit} and across views~\cite{fastvggt2025,goodtokenhunting2026,bui2026cover} act once per input, so here they would keep the same tokens for every target viewpoint.
Visual token pruning in multimodal language models selects image tokens for a query, the analogue of our target viewpoint~\cite{chen2024fastv,zhang2025sparsevlm,sun2026learningpredictmiddlelayerattention,alvar2025divprune,wen2025dart,fang2026prune,zhang2025cdpruner}, and methods that remove redundant tokens find them by comparing features.
Overlapping source views, however, can encode the same surface with dissimilar tokens, so we compare their contributions to the target image.
Sparse DETR~\cite{roh2022sparsedetr} trains a token scorer from its decoder's attention but selects once for all object queries, and sparse attention and KV-cache eviction in language models~\cite{zhang2023h2o,tang2024quest,yang2025tidaldecode}, including selectors distilled from attention~\cite{gao2024seerattention,deepseek2025v32}, score keys with the model's own queries, which for a target viewpoint would require running the decoder.
Our router instead selects tokens per target viewpoint, from the viewpoint alone, for all of its target tokens and decoder blocks.

\paragraph{Selecting source content for a target viewpoint.}
Neural image-based rendering keeps the nearest source views~\cite{wang2021ibrnet,suhail2022gpnr,liu2022neuray} (\cref{sec:exp_selection}) or those overlapping the target viewpoint most under a proxy geometry~\cite{riegler2020fvs}, and Farchione et al.~\cite{farchione2026moregeneticframeselection} learn to choose frames covering a set of target viewpoints.
Camera-controlled video generators retrieve earlier frames overlapping the target viewpoint by field of view~\cite{xiao2025worldmem,yu2025contextasmemory} or visible surfels~\cite{li2025vmem}, or route attention to memory tokens on the same surfaces~\cite{gear2026}.
Classical methods select per scene cell or pixel by a geometric cost of depth, viewing angle and distance~\cite{hedman2016insideout,hedman2018deepblending}, like pixelwise view selection in multi-view stereo~\cite{schoenberger2016pixelwise}; all of these need poses or proxy geometry, the per-pixel ones give every part of the target image the same number of contributions, and whole-view selection flickers when the selection changes~\cite{riegler2021svs}.
Our router selects individual tokens without explicit geometry and learns from the decoder where to spend its budget.

\section{Method}\label{sec:method}
\begin{figure}[t]
\centering
\includegraphics[width=\columnwidth]{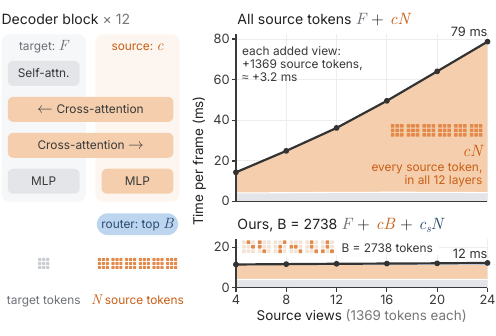}
\caption{\textbf{Where the time to render a viewpoint goes.}
Left: one block of the LagerNVS decoder~\cite{szymanowicz2026lagernvs}.
Gray parts process only target tokens ($F$), orange parts process source tokens ($c$), and blue is our router ($c_s$).
Right: time per frame at $256\times256$ on an RTX PRO 5000, with all source tokens (top) and with our router (bottom); profiled GPU time, scaled to the measured total.}
\label{fig:cost}
\end{figure}

\begin{figure}[t]
\centering
\includegraphics[width=\columnwidth]{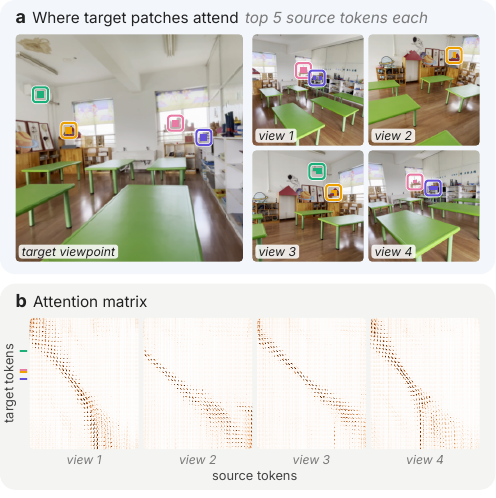}
\caption{\textbf{The decoder attends to few, local source tokens.}
Attention in the decoder block that supervises our router (\cref{sec:method_labels}).
(b)~The 1024 target tokens and the 1369 tokens of each source view are in raster order and max-pooled for display; darker is more attention, and the ticks mark the patches of (a).}
\label{fig:attn}
\end{figure}

\begin{figure*}[t]
\centering
\includegraphics[width=\textwidth]{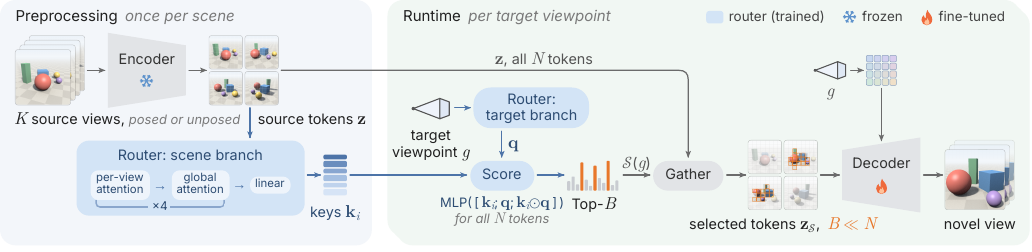}
\caption{\textbf{Overview.}
The source tokens and the keys of the router are computed once per scene and cached, so the scene branch can afford to be heavy.
For every target viewpoint, the lightweight target branch (an MLP) maps the viewpoint to a query, the score head scores all $N$ keys, and the decoder attends only to the $B\ll N$ top-scoring tokens.
The images show a synthetic scene with $8\times8$ tokens per view.}
\label{fig:method}
\end{figure*}

Transformer-based view synthesis models~\cite{jin2025lvsm,szymanowicz2026lagernvs} have two parts.
An encoder turns the $K$ source views into source tokens once per scene, and a decoder takes them and a target viewpoint as input and renders the image seen from that viewpoint.
The decoder represents the target viewpoint by dense Plücker ray embeddings, one per image patch, which form the target tokens, and in each of its $L$ blocks (left of \cref{fig:cost}), the target tokens attend to each other and to the source tokens.
LagerNVS~\cite{szymanowicz2026lagernvs} keeps one source token per patch of every source view, $P=1369$ per view, so its decoder attends to all $N=KP$ source tokens $\mathbf{z}=\{\mathbf{z}_i\}_{i=1}^{N}$.

\paragraph{Where compute is spent.} %
In \cref{fig:cost}, we profile the decoder of LagerNVS~\cite{szymanowicz2026lagernvs} to find where the time to render a viewpoint goes.
The parts that process only target tokens, namely their self-attention and MLP, are gray, and the parts that process source tokens, namely the two cross-attentions between target and source tokens and the MLP of the source tokens, are orange, in the decoder block and the plots alike.
If $F$ is the cost of the layers that process only target tokens (gray) and $c$ the cost of processing one source token through all $L$ blocks (orange), the cost of rendering a viewpoint is
\begin{equation}
  \mathcal{C}_{\text{all}} = \textcolor{costgray}{F} + \textcolor{costorange}{cN},
  \label{eq:cost_all}
\end{equation}

As measured in the top plot of \cref{fig:cost}, the cost of processing the source tokens ($cN$, orange) dominates the time to render a viewpoint and grows with the number of source views: it adds 3.2\,ms with every source view on average and takes 94\% of the 79\,ms needed with 24 source views, whereas $F$ stays near 4\,ms.
We therefore aim to shrink this cost and keep it fixed regardless of the number of source views.
The simplest way is to process only a fixed budget of $B\ll N$ source tokens for every viewpoint.
For a chosen budget $B$, rendering a viewpoint then costs $F+cB$ for any number of source views, much as culling lets a classical renderer rasterize only the primitives that a viewpoint sees.
A complementary direction is to lower $c$, the cost per source token: a decoder that does not update the source tokens can cache their keys and values once per scene~\cite{jia2026efficientlvsm,sun2026dvsm}, but the cost of the source tokens then still grows with $N$.
Compressing the source tokens once per scene, as the encoder-decoder variant of LVSM~\cite{jin2025lvsm} does, fixes their number regardless of the size of the scene, which costs accuracy~\cite{szymanowicz2026lagernvs}.
Classical rendering pipelines instead keep the full representation of a scene and render quickly by discarding what a viewpoint does not see, so we likewise keep all source tokens and select $B$ of them for every viewpoint.

\paragraph{Tokens as local anchors.} 
Our key hypothesis is that, for a given viewpoint, the decoder's attention concentrates on a small fraction of the source tokens (\cref{fig:attn}).
The attention of a target patch falls on a few source tokens that show the same part of the scene in each source view (\cref{fig:attn}a), so that each source token acts as a local anchor in the scene.
This holds for every target token of the viewpoint (\cref{fig:attn}b), and over the 140 test scenes of DL3DV~\cite{ling2024dl3dv}, two views' worth of source tokens receive 85\% of the decoder's attention with 24 source views, while these tokens change from one viewpoint to the next.
This concentration relies on locality: if each source token encoded the whole scene, every token would contribute to every viewpoint, and no fixed budget would suffice.
We therefore give the decoder the source tokens that it attends to most, and with two views' worth of tokens, our model renders within 0.68\,dB of LagerNVS~\cite{szymanowicz2026lagernvs} with all the tokens of 24 source views (\cref{sec:exp_speed}).

\begin{figure}[t]
\centering
\includegraphics[width=\columnwidth]{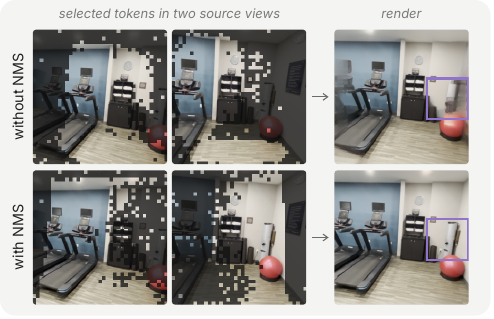}
\caption{\textbf{Non-maximum suppression (NMS) on tokens.} Routers trained without and with NMS select one view's worth of tokens on a target of DL3DV~\cite{ling2024dl3dv}, all from the same 2 of its 8 source views (left; unselected tokens darkened).
The box marks a region that the router without NMS leaves uncovered; \cref{fig:supp_nms_schematic} illustrates NMS on a synthetic scene.}
\label{fig:nms}
\end{figure}

\paragraph{Solution formulation.} The decoder's attention, however, is only available after running the decoder on all source tokens, which is the cost we aim to avoid.
We therefore train a router that predicts this attention from the source tokens and the viewpoint before the decoder runs (\cref{sec:method_router} and \cref{fig:method}), and adapt the pretrained decoder to the sparse subsets of tokens that the router selects (\cref{sec:method_adapt}).
The part of the router that runs for every viewpoint must take little time next to the decoder, or routing would defeat its purpose: most of the router runs once per scene, and the rest takes under 1\,ms of the 12\,ms per frame (bottom plot of \cref{fig:cost}).
View synthesis also relies on redundancy, since each part of the scene must be seen in several source views for its depth and appearance to be estimated.
The source tokens that encode the same part of the scene in these views all receive high attention, so that ranking by attention alone spends the budget on the same parts of the scene several times (\cref{fig:nms}), which we resolve with non-maximum suppression (\cref{sec:method_labels}).

Two simpler ways of choosing the source tokens fall short of our router (\cref{sec:exp_selection}).
Ranking the tokens by a geometric cost~\cite{hedman2018deepblending} computed from depth estimated by MapAnything~\cite{keetha2025mapanything} ignores which tokens the decoder actually uses, and trails our router by 1.62\,dB with 24 source views and one view's worth of tokens.
Giving the decoder whole source views, as image-based renderers select them~\cite{wang2021ibrnet,suhail2022gpnr}, is all or nothing: it renders worse with few source views, and it is unstable along a camera path, since parts of the scene vanish or appear whenever the selected view changes (\cref{fig:view_switch}).
Selecting individual source tokens avoids both problems, since each token is chosen for what it contributes to the viewpoint.

\subsection{Token routing}\label{sec:method_router}

For a target viewpoint $g$, the router scores all $N$ source tokens, and the decoder attends only to the set $\mathcal{S}(g)$ of the $B$ highest-scoring ones (\cref{fig:method}, right).
The cost of rendering a viewpoint from \cref{eq:cost_all} then becomes
\begin{equation}
  \mathcal{C}_{\text{routed}} = \textcolor{costgray}{F} + \textcolor{costorange}{cB} + \textcolor{costblue}{c_s N},
  \label{eq:cost_routed}
\end{equation}
where $c_s$ is the cost of scoring one source token (blue in the bottom plot of \cref{fig:cost}).
The only term that still grows with $N$ is now scaled by $c_s$ instead of $c$, so that the cost depends mainly on the budget $B$ as long as $c_s\ll c$.
To keep $c_s$ small, we split the router into a scene branch and a target branch (\cref{fig:method}).
The scene branch computes for every source token a key that does not depend on the viewpoint, so that it runs only once per scene and can afford to be expensive.
The target branch maps the viewpoint to a query, and a score head compares this query with every key; only these two small parts run for every viewpoint and contribute to $c_s$.

\paragraph{Scene branch.}
The scene branch takes the source tokens as they enter the decoder and applies four pairs of transformer blocks, attending within each source view and then across all source tokens, like the frame and global attention of VGGT~\cite{wang2025vggt}, followed by a linear projection to a key $\mathbf{k}_i\in\mathbb{R}^{256}$ for every token.
Attention across views allows the key of a token to depend on what the other views observe.
The keys are computed once per scene and cached with the source tokens, so the scene branch does not add to the time per frame.

\paragraph{Target branch and scoring.}
The target branch maps the target viewpoint $g$, given as its $3\times4$ camera-to-world matrix in the normalized coordinate frame of the model, to a query $\mathbf{q}(g)=\phi(g)$, where $\phi$ is a three-layer perceptron; for a model that accepts both posed and unposed source views, we add to the query a learned embedding of this mode.
A score head compares every key with the query,
\begin{equation}
  s_i(g) = h\bigl([\mathbf{k}_i;\ \mathbf{q}(g);\ \mathbf{k}_i\odot\mathbf{q}(g)]\bigr),
  \label{eq:score}
\end{equation}
where $h$ is a two-layer MLP and $\odot$ the elementwise product.

\subsection{Router supervision}\label{sec:method_labels}

We train the router to rank the source tokens by how the decoder itself uses them for the target viewpoint, which we read from the decoder's cross-attention (\cref{fig:attn}b).
For every training example, we run the frozen pretrained decoder with all source tokens and take, in one of its intermediate blocks (\cref{sec:supp_details}), the attention probabilities from the target tokens to the source tokens, averaged over heads.
This gives a matrix $A\in\mathbb{R}_{\ge0}^{M\times N}$ over $M$ target patches, from which we define for every source token its attention mass and its cross-attention profile,
\begin{equation}
  a_i = \sum_{p=1}^{M} A_{pi},
  \qquad
  \boldsymbol{\pi}_i = \frac{A_{:,i}}{\lVert A_{:,i}\rVert_2}.
  \label{eq:mass}
\end{equation}
The mass, the attention probability that the target patches assign to the token in total, measures how much the decoder uses the token for the target viewpoint, and the profile shows which part of the target image needs it.

\paragraph{Non-maximum suppression over source tokens.} 
The ranking must account for both the mass and the profile, because a surface that matters for the target viewpoint is usually seen in several source views, and the decoder attends to each of its encodings in the same part of the target image, so their profiles nearly coincide, and ranking by mass alone fills the budget with several encodings of the same surfaces and leaves other parts of the target image uncovered (\cref{fig:nms}).
We account for both with non-maximum suppression~\cite{bodla2017softnms}, which ranks the tokens by mass but discounts the mass of each token by the largest overlap of its profile with those of the tokens already selected, so a token counts for little once the part of the target image that needs it is served by a selected token.
Starting from an empty set $\mathcal{S}$, we repeatedly select the token with the largest gain
\begin{equation}
  \gamma_i(\mathcal{S}) = a_i\Bigl(1-\max_{j\in\mathcal{S}}\,\langle\boldsymbol{\pi}_i,\boldsymbol{\pi}_j\rangle\Bigr)
  \label{eq:gain}
\end{equation}
and record its gain at selection as $\bar\gamma_i$.
Profiles are non-negative unit vectors, so the overlap lies in $[0,1]$, and \cref{eq:gain} adapts the linear decay of Soft-NMS~\cite{bodla2017softnms} with the overlap of profiles in place of that of boxes.
For efficiency, we select several tokens per round before updating the overlaps.

\paragraph{Loss.}
Routing pays off only when the budget is well below $N$, so we run NMS only until it has selected enough tokens for the budgets of interest; these tokens form the head $\mathcal{H}$ of the ranking, and we supervise the router with their recorded gains.
Since these gains never increase along the ranking, the $B$ tokens with the largest gains are exactly the NMS selection for every budget within the head, so a single score per token suffices for all of them.
With the targets
\begin{equation}
  y_i = \log\bigl(\bar\gamma_i+\bar\gamma_{\min}\bigr), \qquad i\in\mathcal{H},
  \label{eq:target}
\end{equation}
where $\bar\gamma_{\min}$ is the smallest recorded gain in the head, we train the router with
\begin{equation}
  \mathcal{L} = 1-\operatorname{corr}_{i\in\mathcal{H}}\bigl(s_i,y_i\bigr)
  + \frac{\lambda}{|\bar{\mathcal{H}}|}\sum_{i\notin\mathcal{H}}\max\bigl(0,\ \hat{s}_i-\hat{s}_f\bigr),
  \label{eq:loss}
\end{equation}
where $\operatorname{corr}$ is the Pearson correlation, which is invariant to the scale and offset of the scores~\cite{huang2022dist}, and the second term pushes every token outside the head below the head token $f$ with the lowest target, with $\hat{s}$ the scores standardized over all $N$ tokens and $\lambda=0.25$.

\subsection{Adapting the decoder}\label{sec:method_adapt}

We take the encoder and the decoder from LagerNVS~\cite{szymanowicz2026lagernvs} and keep the encoder frozen.
The pretrained decoder has only attended to all tokens of every source view, whereas the selected tokens are a sparse subset scattered over parts of many views.
We adapt it to such subsets with low-rank updates~\cite{hu2022lora} to its linear layers, trained with the rendering loss of LagerNVS~\cite{szymanowicz2026lagernvs} on the tokens selected by the frozen router, and we draw the budget at random for every training step, so that a single set of updates serves every budget.
With all source tokens, the updates are switched off, so that our model reduces exactly to LagerNVS~\cite{szymanowicz2026lagernvs}.

\section{Experiments}\label{sec:experiments}
\providecommand{\tbd}{\textcolor{red}{?}}
\providecommand{\refrow}[1]{\textcolor{gray}{#1}}

We first measure how routing trades speed for quality as the number of source views grows (\cref{sec:exp_speed}), then compare our model with feed-forward methods with and without input poses (\cref{sec:exp_comparison}), and analyze its token selection (\cref{sec:exp_selection}).

\subsection{Experimental setup}\label{sec:exp_setup}

\paragraph{Models.}
We apply our method to two models of LagerNVS~\cite{szymanowicz2026lagernvs}: the model trained on DL3DV~\cite{ling2024dl3dv}, which renders $256\times256$ views from 2 to 6 posed source views, and the general model, which renders $512\times512$ views from source views with or without poses.
For each, we train the router and adapt the decoder with the training protocol of LagerNVS (\cref{sec:supp_details}), and with all tokens ($B=N$) our model reduces exactly to its base model.

\paragraph{Benchmarks.}
With input poses, we evaluate on the 140 scenes of the DL3DV benchmark~\cite{ling2024dl3dv}, with 4 to 24 source views spread over the first 300 frames of each scene, and on its splits of DepthSplat~\cite{xu2025depthsplat} and ReSplat~\cite{xu2026resplat}, with 4 to 32 source views.
Without input poses, we evaluate on RealEstate10K~\cite{zhou2018stereo} with 2 source views, on the split of NoPoSplat~\cite{ye2025noposplat}, and on CO3D~\cite{reizenstein2021co3d} and Mip-NeRF 360~\cite{barron2022mipnerf360} with 9 source views, on the splits of ReconFusion~\cite{wu2024reconfusion}.
\Cref{sec:supp_details} gives the source views and target viewpoints of each benchmark.

\paragraph{Baselines and metrics.}
With input poses, we compare with the feed-forward Gaussian splatting methods DepthSplat~\cite{xu2025depthsplat} and ReSplat~\cite{xu2026resplat}, and with CLiFT~\cite{wang2025clift}, which condenses the source tokens of each scene into fewer tokens, all with their public models and the wider inputs of their training.
Without input poses, we compare with NoPoSplat~\cite{ye2025noposplat}, SPFSplatV2~\cite{huang2026spfsplatv2}, YoNoSplat~\cite{ye2026yonosplat} and WorldMirror~\cite{liu2025worldmirror}.
We report PSNR, SSIM and LPIPS, on the central square of each frame for DL3DV~\cite{ling2024dl3dv}.

\subsection{Speed}\label{sec:exp_speed}
\begin{figure}[t]
\centering
\includegraphics[width=\columnwidth]{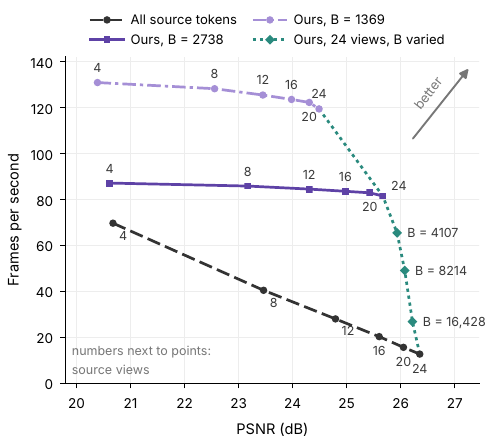}
\caption{\textbf{Speed and quality.}
Frames per second against PSNR on the 140 test scenes of DL3DV~\cite{ling2024dl3dv}, with 4 to 24 source views (numbers next to the points) spaced evenly over the first 300 frames of each scene and the same 228 target viewpoints per scene at every number of views.
$B$ is the number of source tokens, 1369 per view, that our decoder attends to per frame; teal shows increasing $B$ at 24 views, up to all 32,856 tokens.
Speed is measured at $256\times256$ on an RTX PRO 5000 and excludes the work done once per scene, such as encoding the source views (\cref{sec:supp_speed}).}
\label{fig:speed}
\end{figure}
In \cref{fig:speed}, a better renderer lies further toward the upper right, with a higher PSNR at a higher frame rate.
Routing moves LagerNVS~\cite{szymanowicz2026lagernvs} in this direction: with two views' worth of tokens, our model renders 82 to 87 frames per second at every number of source views, against 13 to 70 for LagerNVS with all source tokens, with a PSNR at most 0.68\,dB lower.
To raise its quality, LagerNVS has to add source views, but every view adds tokens that its decoder attends to in every frame, so going from 4 to 24 views raises its PSNR by 5.7\,dB but makes it 5.5 times slower.
With a fixed budget, our model gains 5.1\,dB from the same views while its frame rate drops by only 6\%, since scoring a source token is far cheaper than attending to it.
Once the source views are given, the budget is also a better way than the number of views to trade quality for speed, as the teal curve of our model, which varies the budget at 24 views, lies above the black curve of LagerNVS, which varies the number of views: to render more than three times as fast as with all tokens of 24 views, our model lowers its budget to a quarter of the tokens and loses 0.28\,dB, whereas LagerNVS has to drop to 8 views and loses 2.89\,dB.
Since a single ranking serves every budget, this choice can change from one frame to the next without encoding the scene again.

\subsection{Comparison with prior methods}\label{sec:exp_comparison}
\begin{table}[t]
\centering
\caption{Comparison without input poses on RealEstate10K~\cite{zhou2018stereo} (NoPoSplat~\cite{ye2025noposplat} split, scored at $256\times256$), CO3D~\cite{reizenstein2021co3d} and Mip-NeRF 360~\cite{barron2022mipnerf360} (ReconFusion~\cite{wu2024reconfusion} splits, scored at $512\times512$).
Every method estimates and refines the target pose.
$^*$Numbers from the papers. $^\dagger$Trained on CO3D.
Ours builds on the general model of LagerNVS~\cite{szymanowicz2026lagernvs} and reduces to it with all tokens (gray); 1369 tokens are one view.
FPS counts target viewpoints rendered per second on one RTX PRO 5000 once the scene is encoded, with the fastest renderer that reproduces each splatting method's images, at the refined target pose and excluding the refinement (\cref{tab:supp_speed_cmp}).
Colors mark the \rankkey{} entry in each quality column of each block, excluding the gray rows.}
\label{tab:cmp_posefree}
\footnotesize
\setlength{\tabcolsep}{3pt}
\renewcommand{\arraystretch}{1.1}
\begin{tabular}{@{}llcccc@{}}
\toprule
Method & Test data & PSNR$\uparrow$ & SSIM$\uparrow$ & LPIPS$\downarrow$ & FPS$\uparrow$\\
\midrule
NoPoSplat$^*$~\cite{ye2025noposplat} & \multirow{6}{*}{\shortstack[l]{RE10K\\2 views}} & \trd 25.03 & \trd 0.838 & \trd 0.160 & 868\\
SPFSplatV2$^*$~\cite{huang2026spfsplatv2} & & \snd 26.16 & \snd 0.861 & \snd 0.146 & 828\\
YoNoSplat~\cite{ye2026yonosplat} & & 24.63 & 0.824 & 0.176 & 867\\ %
WorldMirror~\cite{liu2025worldmirror} & & 22.74 & 0.792 & 0.208 & 847\\
Ours, $B{=}1369$ & & \fst 27.27 & \fst 0.865 & \fst 0.142 & 41\\
\refrow{LagerNVS~\cite{szymanowicz2026lagernvs}, $B{=}N$} & & \refrow{27.61} & \refrow{0.872} & \refrow{0.136} & \refrow{34}\\
\midrule
YoNoSplat~\cite{ye2026yonosplat} & \multirow{6}{*}{\shortstack[l]{CO3D\\9 views}} & 22.08 & 0.715 & 0.388 & 493\\
WorldMirror$^\dagger$~\cite{liu2025worldmirror} & & 21.12 & 0.712 & 0.393 & 527\\
Ours, $B{=}1369$ & & \trd 25.45 & \trd 0.775 & \trd 0.297 & 42\\
Ours, $B{=}2738$ & & \snd 26.04 & \snd 0.784 & \snd 0.282 & 33\\
Ours, $B{=}4107$ & & \fst 26.18 & \fst 0.787 & \fst 0.278 & 29\\
\refrow{LagerNVS~\cite{szymanowicz2026lagernvs}, $B{=}N$} & & \refrow{26.43} & \refrow{0.793} & \refrow{0.271} & \refrow{17}\\
\midrule
YoNoSplat~\cite{ye2026yonosplat} & \multirow{6}{*}{\shortstack[l]{Mip-NeRF\\360, 9 views}} & 18.67 & 0.439 & 0.525 & 504\\
WorldMirror~\cite{liu2025worldmirror} & & 17.71 & 0.418 & 0.502 & 498\\
Ours, $B{=}1369$ & & \trd 19.99 & \trd 0.484 & \trd 0.426 & 42\\
Ours, $B{=}2738$ & & \snd 20.32 & \snd 0.494 & \snd 0.410 & 33\\
Ours, $B{=}4107$ & & \fst 20.40 & \fst 0.496 & \fst 0.407 & 29\\ %
\refrow{LagerNVS~\cite{szymanowicz2026lagernvs}, $B{=}N$} & & \refrow{20.65} & \refrow{0.503} & \refrow{0.402} & \refrow{18}\\ %
\bottomrule
\end{tabular}
\end{table}

\paragraph{Without input poses.}
With one view's worth of tokens, our model outperforms every pose-free baseline on all three benchmarks, the strongest by 1.1 to 3.4\,dB in PSNR, and with three views' worth of the nine on CO3D and Mip-NeRF 360, it stays within 0.25\,dB of LagerNVS~\cite{szymanowicz2026lagernvs} (\cref{tab:cmp_posefree}).
Following the protocol of NoPoSplat~\cite{ye2025noposplat}, every method in \cref{tab:cmp_posefree} estimates the target pose and refines it against the target image before scoring, a step needed only for evaluation, and our router selects its tokens once from the initial pose.

\paragraph{With input poses.}
On DL3DV (\cref{tab:supp_cmp_depthsplat,tab:supp_cmp_resplat}), one view's worth of tokens outperforms DepthSplat~\cite{xu2025depthsplat} and CLiFT~\cite{wang2025clift} by over 3\,dB with 4 and 6 views.
Our model outperforms ReSplat~\cite{xu2026resplat} at 8 views with any budget and at 16 views from two views' worth, with lower LPIPS throughout, and trails it by 0.1 to 0.3\,dB at 32 views, where ReSplat falls to 27.60\,dB on our crops.

\paragraph{Frame rate.}
Once their Gaussians are predicted, the splatting methods render a frame 3 to 21 times faster than our model with one view's worth of tokens, since rasterizing Gaussians costs far less than running a transformer decoder, whereas ours renders 1.2 to 12 times faster than LagerNVS~\cite{szymanowicz2026lagernvs}, with the largest gains for the most source views (\cref{tab:cmp_posefree} and \cref{tab:supp_speed_cmp}).

\subsection{Selection strategies}\label{sec:exp_selection}
\begin{figure}[t]
\centering
\includegraphics[width=\columnwidth]{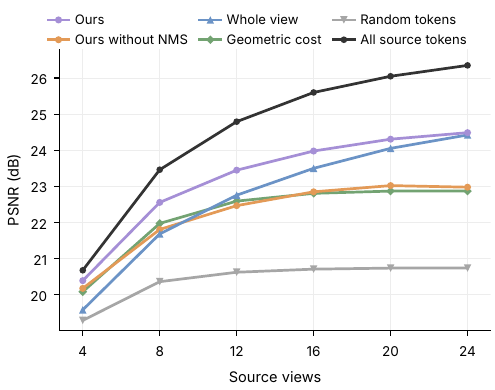}
\caption{\textbf{Selection strategies.}
PSNR with one view's worth of source tokens ($B=1369$) on the 140 test scenes of DL3DV~\cite{ling2024dl3dv}, with 4 to 24 source views spaced evenly over the first 300 frames and the same 228 target viewpoints per scene.
Each strategy renders through a decoder adapted to its own selections in the same way as ours, and LagerNVS~\cite{szymanowicz2026lagernvs} attends to all source tokens.}
\label{fig:selection}
\end{figure}
\definecolor{attnc}{HTML}{E8772A}
\definecolor{oursc}{HTML}{2F6DB5}

\begin{figure}[t]
\centering
\includegraphics[width=\columnwidth]{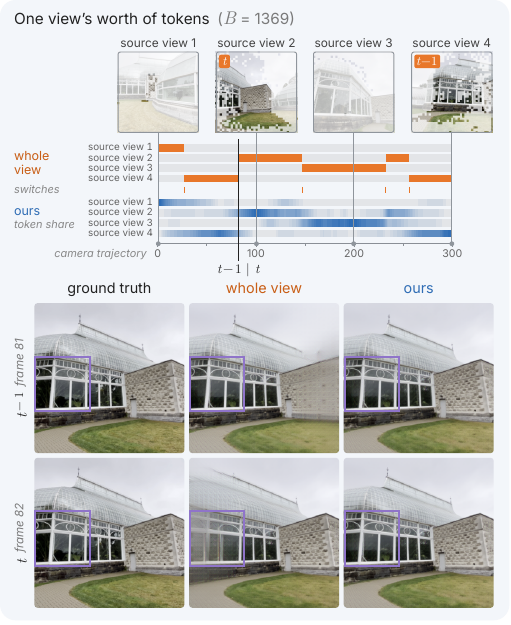}
\caption{Switching source views along a 300-frame trajectory, with four source views captured at the vertical lines.
The whole view (\textcolor{attnc}{orange}) is the source view from which our router selects the most tokens, and tags mark the views it uses at $t{-}1$ and $t$; \textcolor{oursc}{blue} lanes show the share of our router's tokens from each view, and the views show its tokens at frame $t$.
At the marked switch, the boxed content turns translucent with the whole view, while our router keeps 95\% of its tokens.
\Cref{fig:supp_view_switch_1369,fig:supp_view_switch_2738} show further switches, also with two views' worth of tokens.}
\label{fig:view_switch}
\end{figure}

\Cref{fig:selection} compares our router with other ways of choosing the source tokens, each rendered through a decoder adapted to its own selections in the same way as ours, with one view's worth of tokens.
The gap between our model and LagerNVS~\cite{szymanowicz2026lagernvs} with all tokens grows with the number of views, from 0.29 to 1.86\,dB, since this budget is an ever smaller share of the source tokens, down to 4\% with 24 views, but doubling the budget to two views' worth already closes it to 0.68\,dB (\cref{fig:speed}).

The other strategies in \cref{fig:selection} choose tokens as follows:
\begin{itemize}[leftmargin=*,topsep=2pt,itemsep=1pt]
\item \textbf{Ours without NMS} uses a router trained to rank the tokens by their attention mass alone (\cref{eq:mass}).
\item \textbf{Whole view} gives the decoder all tokens of the source view from which our router selects the most tokens.
It renders 0.1 to 0.2\,dB better than the source view nearest to the target viewpoint, which the appendix compares as well.
Image-based renderers~\cite{wang2021ibrnet,suhail2022gpnr} select their source views by such proximity in pose, as do generative view synthesis models when they retrieve views from a memory~\cite{zhou2025seva}; we measure it with the distance of GP-MVS~\cite{hou2019gpmvs}, which combines translation and rotation.
\item \textbf{Geometric cost} ranks the tokens by the per-pixel cost that Deep Blending~\cite{hedman2018deepblending} computes from depth, viewing angle and distance, with depth from MapAnything~\cite{keetha2025mapanything}.
\item \textbf{Random tokens} are drawn uniformly per viewpoint.
\end{itemize}

Our router outperforms all of them with 4 views; as views are added, the gap to the whole view closes, from 0.81 to 0.06\,dB with 24 views, where the two are on par, since denser source views make it more likely that one nearly coincides with the target viewpoint.
The gap to our model without NMS instead grows, from 0.21 to 1.51\,dB, since denser source views encode the same surfaces several times, and ranking by attention mass alone spends the budget on them.
The geometric cost comes closest to our router with 4 views, 0.31\,dB below it, but levels off from 16 views, as random tokens do from 12, and trails it by 1.62\,dB with 24 views.

Whole views also make the rendering unstable along a camera path, since each change of the selected view replaces all the tokens that the decoder sees, so that parts of the scene vanish, turn translucent or appear from one frame to the next (\cref{fig:view_switch}).
With four source views, the whole view changes 8 times over 300 frames on average, and at these switches its PSNR changes by 1.51\,dB from one frame to the next, about four times as much as between other frames.
Our router keeps 93\% of its tokens at these switches, and its PSNR changes by only 0.28\,dB there, as much as that of LagerNVS~\cite{szymanowicz2026lagernvs} with all tokens.
The appendix repeats \cref{fig:selection} with two views' worth of tokens and adds the nearest source view and the selection rule of CLiFT~\cite{wang2025clift} (\cref{sec:supp_selection}).

\section{Discussion}\label{sec:discussion}
Classical rendering pipelines have long rasterized only what a viewpoint sees, and our results suggest that transformer-based view synthesis models can learn to do the same from their own attention.
To achieve this,  we use the intuition that each source token acts as a local anchor, similar to a mesh element, and hence, each target token only needs to attend to a small set of source tokens.
We design a token routing scheme that keeps the quality of transformer-based view synthesis models while their rendering time per viewpoint barely grows with the number of source views.

\paragraph{Hardware with little compute.}
On hardware with little compute bandwidth, transformer-based view synthesis models remain a poor fit even with token routing, given their current size.
Below one view's worth of tokens, the quality of our model drops quickly (\cref{sec:supp_budgets}), while its rendering time barely decreases, because every target token still passes through the full decoder.
Complementary reductions, such as smaller distilled decoders, lower numerical precision or fewer target tokens, may eventually make these models practical on such hardware.

\paragraph{Streaming operation.} The issue of compute bandwidth gets all the more important for streaming applications, where the encoding needs to happen in real time as well.
Token routing only speeds up rendering, not the encoding of the source views, whose time grows with their number.
Here, ideas from streaming geometry transformers~\cite{ICLR2026_8da04a60,stream3r2025,cut3r} could make the encoders of view synthesis models faster and let them take in source views as they arrive.

\paragraph{Pruning or merging.}
Our router prunes source tokens rather than merging them, a choice on which studies of token reduction in image classification disagree~\cite{bolya2023tome,Haurum_2023_ICCVW,kim2024token}.
Pruning suits our setting because one ranking of the tokens serves every budget, so the budget can change at render time, even between frames.
Yet classical approaches to content synthesis avoid redundancy in their representations by merging scans of an object or scene into a single consistent mesh.
Our approach would benefit from such merging, which reduces the number of source tokens to prune.

\paragraph{Multiscale representations.} Another limitation of transformer-based rendering is that we have no graceful approach to handling resolution/scale. Consider a large scene rendered from a distant viewpoint; the lack of scale implies that we will attend to a large collection of source tokens whose contribution  in the target image could even be smaller than a pixel. All of this necessitates a broader range of operations for token-based representations, akin to what has been developed for meshes.

\section*{LLM Usage Statement}
We used Claude (Anthropic) throughout this project.
The authors drafted the introduction, which Claude edited, and Claude generated drafts of the other sections and summaries of related work, which the authors reviewed and edited.
Claude also generated code for training, evaluation, timing, plotting and video rendering, as well as figure layouts, all of which the authors reviewed and approved.
For the experiments, the data and the research ideas, Claude only assisted by editing or refining the authors' own work.
Every citation was checked against the original paper, and all BibTeX entries were taken from official sources.
The authors made all main decisions on the method and the experiments, verified all code, numbers and figures, and take full responsibility for the content of the paper.

\nocite{szymanowicz2026lagernvs}
{
    \small
    \bibliographystyle{ieeenat_fullname}
    \bibliography{references}
}

\clearpage
\appendix
\setcounter{figure}{0}\setcounter{table}{0}\setcounter{equation}{0}
\renewcommand{\thefigure}{S\arabic{figure}}
\renewcommand{\thetable}{S\arabic{table}}
\renewcommand{\theequation}{S\arabic{equation}}
\renewcommand{\theHfigure}{S\arabic{figure}}
\renewcommand{\theHtable}{S\arabic{table}}
\renewcommand{\theHequation}{S\arabic{equation}}

\section{Overview}\label{sec:supp_overview}
This appendix gives the details of our experiments and additional results.
\Cref{sec:supp_details} describes the training, the benchmarks, the inputs of the baselines and the pose-free protocol, and \cref{sec:supp_full} gives our comparison with input poses and the full version of \cref{tab:cmp_posefree}.
\Cref{sec:supp_resplat} compares the publicly available models of ReSplat~\cite{xu2026resplat}, \cref{sec:supp_finetune} evaluates our model without fine-tuning at the test view count, \cref{sec:supp_budgets} reports budgets below one view's worth of tokens, \cref{sec:supp_posefree} adds further pose-free methods and view counts, \cref{sec:supp_nms} illustrates our non-maximum suppression, \cref{sec:supp_labelblock} compares the decoder blocks from which the labels can be computed, \cref{sec:supp_selection} adds further selection strategies, \cref{sec:supp_view_switch} shows further view switches, and \cref{sec:supp_speed} gives the measurements behind the speed analysis, with timings at $512\times512$.
\Cref{sec:supp_posed_refine} repeats the comparison with input poses on the ReSplat split with the target pose refined as in the pose-free protocol.

\section{Experimental details}\label{sec:supp_details}

\paragraph{Models.}
As in \cref{sec:exp_setup}, we apply our method to two released models of LagerNVS~\cite{szymanowicz2026lagernvs}, with a router and a decoder adaptation of its own for each.
The model trained on DL3DV~\cite{ling2024dl3dv} renders $256\times256$ views from 2 to 6 posed source views, and all our experiments with input poses build on it: the speed and selection experiments of \cref{sec:exp_speed,sec:exp_selection}, the comparisons on DL3DV (\cref{sec:supp_full,sec:supp_finetune,sec:supp_budgets,sec:supp_posed_refine}) and \cref{sec:supp_nms,sec:supp_labelblock,sec:supp_selection,sec:supp_view_switch}.
The general model renders $512\times512$ views from 1 to 10 source views with or without their poses, and we use it only for the comparison without input poses, in \cref{tab:cmp_posefree,tab:supp_full_re10k,tab:supp_full_co3d,tab:supp_full_mip360} and \cref{sec:supp_posefree}, and for the timing at $512\times512$ in \cref{tab:supp_speed512}.

\paragraph{Training.}
For the model of LagerNVS~\cite{szymanowicz2026lagernvs} trained on DL3DV~\cite{ling2024dl3dv}, we train the router from scratch for 100k iterations, with labels computed from the sixth of the twelve decoder blocks, which \cref{sec:supp_labelblock} compares with other blocks.
We then adapt the decoder for 100k iterations with rank-32 low-rank updates~\cite{hu2022lora} to all of its linear layers, which amount to 12.6\,M trainable parameters, or 6.6\% of the decoder.
During adaptation, the budget is drawn log-uniformly between half a view (685 tokens) and all $N$ tokens, and one step in ten uses all tokens.
Both stages are trained on DL3DV-10K~\cite{ling2024dl3dv} with the training protocol of LagerNVS, with 2 to 6 source views and eight target viewpoints per scene, and take about 1.3 and 1.4 days on eight H200 GPUs.
For the general model of LagerNVS, we train the router and adapt the decoder in the same way, for 100k iterations each, with the training protocol of that model, which samples 1 to 10 source views with or without their poses.
With all tokens ($B=N$), the updates are not applied, so that our model reduces exactly to LagerNVS.

\paragraph{Benchmarks.}
The DepthSplat split~\cite{xu2025depthsplat} places 4 or 6 source views within the first 50 frames of each of the 140 scenes of the DL3DV benchmark~\cite{ling2024dl3dv}, and, as in LagerNVS~\cite{szymanowicz2026lagernvs}, we render the remaining 46 or 44 frames of this window, so that no source view is also a target.
The ReSplat split~\cite{xu2026resplat} uses 8, 16 or 32 source views that span 40, 80 or 160 frames, with 8, 16 or 24 target viewpoints per scene that are disjoint from the sources.
Its splits for different view counts contain different targets and differ in difficulty, so results are comparable only within a view count.
For the experiments on speed and selection, we place $K$ source views at frames $\mathrm{round}(299\,k/(K-1))$, $k=0,\dots,K-1$, of the first 300 frames of each scene, for $K$ from 4 to 24, and render the same 228 target viewpoints per scene at every $K$, namely the frames that are a source view at no $K$.
Without input poses, RealEstate10K~\cite{zhou2018stereo} uses the 5,601 test scenes of the split of NoPoSplat~\cite{ye2025noposplat}, scored at $256\times256$ with its protocol and metric code, and CO3D~\cite{reizenstein2021co3d} and Mip-NeRF 360~\cite{barron2022mipnerf360} use the 20 sequences and 9 scenes of the splits of ReconFusion~\cite{wu2024reconfusion}, scored at $512\times512$ with the metric code of LagerNVS.

\paragraph{Inputs of the baselines.}
Like LagerNVS~\cite{szymanowicz2026lagernvs}, our model operates on the central $540\times540$ square of each $960\times540$ frame of DL3DV~\cite{ling2024dl3dv}, resized to $256\times256$, and we score all methods on these crops.
DepthSplat~\cite{xu2025depthsplat}, ReSplat~\cite{xu2026resplat} and CLiFT~\cite{wang2025clift} are trained on $256\times448$ inputs taken from $270\times480$ versions of the same frames.
DepthSplat crops its inputs from these frames, so that they cover 1.66 times the width and 0.95 times the height of our crops, while ReSplat and CLiFT resize the frames, so that their inputs cover 1.75 times the width at the same height. %
In the main text, the baselines receive these wider inputs, with which they score higher, and \cref{tab:supp_cmp_depthsplat,tab:supp_cmp_resplat} also give DepthSplat and ReSplat the same square crops as our model.

\paragraph{Baselines.}
For DepthSplat~\cite{xu2025depthsplat}, we use its model trained on DL3DV~\cite{ling2024dl3dv} with 2 to 6 source views.
ReSplat~\cite{xu2026resplat} provides a separate model for each view count, and we use its base model with four refinement iterations, its strongest available configuration on this benchmark (\cref{sec:supp_resplat}).
For CLiFT~\cite{wang2025clift}, we use its model trained on DL3DV with 4 to 6 source views, with one cluster per source token, so that it keeps all of its tokens, its highest-quality setting.
Since the release of DL3DV from which CLiFT reads its frames is incomplete for 8 of the benchmark scenes, we give it the same frames and cameras as every other method, downsampled to the $270\times480$ frames of its training and resized by its own code. %
Under their original evaluation protocols, DepthSplat and ReSplat reproduce the results reported by their authors to within 0.07\,dB, and CLiFT, with the frames of the release, matches its published results on DL3DV, which are given only as plots, to within about 0.1\,dB. %
Because each model of ReSplat is trained for a single view count, on the ReSplat split we fine-tune our router and decoder at each view count, for 2k iterations each, with the view sampling used to train ReSplat and our square crops (\cref{sec:supp_finetune}).

\paragraph{Pose-free protocol.}
Following NoPoSplat~\cite{ye2025noposplat}, every method estimates the target pose from the target image and refines it with 100 Adam steps at a learning rate of 0.005 on the mean squared error plus 0.05 times the LPIPS between the rendered and the target image.
Our router selects the tokens once from the initial target pose, and the selection is held fixed during refinement, which changes the PSNR by at most 0.06\,dB compared with selecting the tokens again at every step.
For YoNoSplat~\cite{ye2026yonosplat}, we report the stronger of its own test-time alignment and this refinement, and WorldMirror~\cite{liu2025worldmirror} starts the refinement from its own predicted target pose; both receive the ground-truth intrinsics.
The numbers of NoPoSplat and SPFSplatV2~\cite{huang2026spfsplatv2} are taken from their papers.

\section{Full comparison tables}\label{sec:supp_full}
\Cref{tab:supp_cmp_depthsplat,tab:supp_cmp_resplat} give our comparison with input poses, which \cref{sec:exp_comparison} summarizes, with DepthSplat~\cite{xu2025depthsplat} and ReSplat~\cite{xu2026resplat} also on the square crops of our model, and \cref{tab:supp_full_re10k,tab:supp_full_co3d,tab:supp_full_mip360} are the full versions of \cref{tab:cmp_posefree}, with all three budgets.
\Cref{tab:supp_posed_refine} in \cref{sec:supp_posed_refine} repeats \cref{tab:supp_cmp_resplat} with the target pose refined.

\begin{table}[t]
\centering
\caption{Comparison on the DepthSplat split~\cite{xu2025depthsplat} of DL3DV~\cite{ling2024dl3dv} (140 scenes).
All methods are scored on the central square of each frame, resized to $256\times256$.
Subscripts give the input resolution: DepthSplat~\cite{xu2025depthsplat} receives either the same $256\times256$ square crops as our model or $256\times448$ inputs as in its training, and CLiFT~\cite{wang2025clift} receives $256\times448$ inputs only.
\textbf{\boldmath$^*$These inputs cover $1.66\times$ (DepthSplat~\cite{xu2025depthsplat}, at $0.95\times$ the height) and $1.75\times$ (CLiFT~\cite{wang2025clift}) the width of our crops.}
Ours selects $B$ source tokens per target viewpoint (1369 tokens are one view); with all $N$ tokens it reduces to LagerNVS~\cite{szymanowicz2026lagernvs} (gray).
Colors mark the \rankkey{} entry in each column, excluding the gray row.}
\label{tab:supp_cmp_depthsplat}
\scriptsize
\setlength{\tabcolsep}{1.2pt}
\renewcommand{\arraystretch}{1.1}
\begin{tabular}{@{}lcccccc@{}}
\toprule
& \multicolumn{3}{c}{4 views} & \multicolumn{3}{c}{6 views}\\
\cmidrule(lr){2-4}\cmidrule(l){5-7}
Method & PSNR$\uparrow$ & SSIM$\uparrow$ & LPIPS$\downarrow$ & PSNR$\uparrow$ & SSIM$\uparrow$ & LPIPS$\downarrow$\\
\midrule
DepthSplat$_{256\times256}$~\cite{xu2025depthsplat} & 22.39 & 0.767 & 0.186 & 23.56 & 0.813 & 0.151\\
DepthSplat$_{256\times448}^{*}$~\cite{xu2025depthsplat} & 23.06 & 0.786 & 0.172 & 23.87 & 0.821 & 0.147\\
CLiFT$_{256\times448}^{*}$~\cite{wang2025clift} & 23.42 & 0.754 & 0.203 & 24.78 & 0.798 & 0.166\\ %
\midrule
Ours$_{256\times256}$, $B{=}1369$ & \trd 26.79 & \trd 0.857 & \trd 0.105 & \trd 28.22 & \trd 0.889 & \trd 0.082\\ %
Ours$_{256\times256}$, $B{=}2738$ & \snd 27.41 & \snd 0.867 & \snd 0.096 & \snd 29.18 & \snd 0.901 & \snd 0.070\\
Ours$_{256\times256}$, $B{=}4107$ & \fst 27.48 & \fst 0.869 & \fst 0.095 & \fst 29.28 & \fst 0.903 & \fst 0.069\\ %
\midrule
\refrow{LagerNVS$_{256\times256}$~\cite{szymanowicz2026lagernvs}, $B{=}N$} & \refrow{27.60} & \refrow{0.871} & \refrow{0.094} & \refrow{29.49} & \refrow{0.906} & \refrow{0.068}\\
\bottomrule
\end{tabular}
\end{table}

\begin{table*}[t]
\centering
\caption{Comparison on the ReSplat split~\cite{xu2026resplat} of DL3DV~\cite{ling2024dl3dv} (140 scenes).
ReSplat~\cite{xu2026resplat} uses a separate model for each view count, and on this split we also fine-tune ours at each view count (2k iterations).
Notation as in \cref{tab:supp_cmp_depthsplat}.
\textbf{\boldmath$^*$Inputs of $256\times448$ pixels, as in training, cover $1.75\times$ the width of our crops.}
$^\dagger$LagerNVS~\cite{szymanowicz2026lagernvs} was trained with 2 to 6 source views and is not fine-tuned here.
The three view counts use different target viewpoints, so numbers are comparable only within a view count.
Colors mark the \rankkey{} entry in each column, excluding the gray row.}
\label{tab:supp_cmp_resplat}
\footnotesize
\setlength{\tabcolsep}{6pt}
\renewcommand{\arraystretch}{1.1}
\begin{tabular}{@{}lccccccccc@{}}
\toprule
& \multicolumn{3}{c}{8 views} & \multicolumn{3}{c}{16 views} & \multicolumn{3}{c}{32 views}\\
\cmidrule(lr){2-4}\cmidrule(lr){5-7}\cmidrule(l){8-10}
Method & PSNR$\uparrow$ & SSIM$\uparrow$ & LPIPS$\downarrow$ & PSNR$\uparrow$ & SSIM$\uparrow$ & LPIPS$\downarrow$ & PSNR$\uparrow$ & SSIM$\uparrow$ & LPIPS$\downarrow$\\
\midrule
ReSplat$_{256\times256}$~\cite{xu2026resplat} & 27.78 & 0.887 & 0.129 & 28.04 & 0.889 & 0.124 & 27.60 & \snd 0.882 & 0.130\\
ReSplat$_{256\times448}^{*}$~\cite{xu2026resplat} & 29.14 & 0.911 & 0.090 & \trd 28.98 & \trd 0.908 & 0.092 & \fst 28.31 & \fst 0.900 & 0.101\\ %
\midrule
Ours$_{256\times256}$, $B{=}1369$ & \trd 30.27 & \trd 0.923 & \trd 0.058 & 28.68 & 0.896 & \trd 0.072 & 26.91 & 0.856 & \trd 0.090\\
Ours$_{256\times256}$, $B{=}2738$ & \snd 31.37 & \snd 0.935 & \snd 0.048 & \snd 29.86 & \snd 0.911 & \snd 0.058 & \trd 28.05 & 0.877 & \snd 0.072\\ %
Ours$_{256\times256}$, $B{=}4107$ & \fst 31.51 & \fst 0.936 & \fst 0.047 & \fst 30.00 & \fst 0.913 & \fst 0.057 & \snd 28.19 & \trd 0.880 & \fst 0.071\\ %
\midrule
\refrow{LagerNVS$^\dagger_{256\times256}$~\cite{szymanowicz2026lagernvs}, $B{=}N$} & \refrow{31.77} & \refrow{0.938} & \refrow{0.046} & \refrow{30.20} & \refrow{0.915} & \refrow{0.056} & \refrow{27.83} & \refrow{0.868} & \refrow{0.074}\\
\bottomrule
\end{tabular}
\end{table*}

\begin{table}[t]
\centering
\caption{Comparison with pose-free methods on RealEstate10K~\cite{zhou2018stereo} with 2 source views (NoPoSplat split, 5,601 scenes), scored at $256\times256$ with the protocol and metric code of NoPoSplat~\cite{ye2025noposplat}.
Every method estimates the target pose from the target image and refines it on the rendered image; the numbers of NoPoSplat~\cite{ye2025noposplat} and SPFSplatV2~\cite{huang2026spfsplatv2} are from their papers.
Ours selects $B$ of the $N{=}2738$ source tokens; with all tokens it reduces to LagerNVS~\cite{szymanowicz2026lagernvs} (gray).
Colors mark the \rankkey{} entry in each column, excluding the gray row.}
\label{tab:supp_full_re10k}
\footnotesize
\setlength{\tabcolsep}{5pt}
\renewcommand{\arraystretch}{1.1}
\begin{tabular}{@{}lccc@{}}
\toprule
Method & PSNR$\uparrow$ & SSIM$\uparrow$ & LPIPS$\downarrow$\\
\midrule
NoPoSplat~\cite{ye2025noposplat} & \trd 25.03 & \trd 0.838 & \trd 0.160\\
SPFSplatV2~\cite{huang2026spfsplatv2} & \snd 26.16 & \snd 0.861 & \snd 0.146\\
YoNoSplat~\cite{ye2026yonosplat} & 24.63 & 0.824 & 0.176\\ %
WorldMirror~\cite{liu2025worldmirror} & 22.74 & 0.792 & 0.208\\ %
\midrule
Ours, $B{=}1369$ & \fst 27.27 & \fst 0.865 & \fst 0.142\\
\midrule
\refrow{LagerNVS~\cite{szymanowicz2026lagernvs}, $B{=}N$} & \refrow{27.61} & \refrow{0.872} & \refrow{0.136}\\
\bottomrule
\end{tabular}
\end{table}

\begin{table}[t]
\centering
\caption{Comparison with pose-free methods on CO3D~\cite{reizenstein2021co3d} with 9 source views (ReconFusion split~\cite{wu2024reconfusion}, 20 sequences), scored at $512\times512$ with the metric code of LagerNVS~\cite{szymanowicz2026lagernvs}.
Every method estimates the target pose from the target image and refines it on the rendered image.
$^\dagger$Trained on CO3D~\cite{reizenstein2021co3d}.
Ours selects $B$ of the $N{=}12{,}321$ source tokens (1369 tokens are one view); with all tokens it reduces to LagerNVS~\cite{szymanowicz2026lagernvs} (gray).
Colors mark the \rankkey{} entry in each column, excluding the gray row.}
\label{tab:supp_full_co3d}
\footnotesize
\setlength{\tabcolsep}{5pt}
\renewcommand{\arraystretch}{1.1}
\begin{tabular}{@{}lccc@{}}
\toprule
Method & PSNR$\uparrow$ & SSIM$\uparrow$ & LPIPS$\downarrow$\\
\midrule
YoNoSplat~\cite{ye2026yonosplat} & 22.08 & 0.715 & 0.388\\ %
WorldMirror$^\dagger$~\cite{liu2025worldmirror} & 21.12 & 0.712 & 0.393\\ %
\midrule
Ours, $B{=}1369$ & \trd 25.45 & \trd 0.775 & \trd 0.297\\
Ours, $B{=}2738$ & \snd 26.04 & \snd 0.784 & \snd 0.282\\
Ours, $B{=}4107$ & \fst 26.18 & \fst 0.787 & \fst 0.278\\
\midrule
\refrow{LagerNVS~\cite{szymanowicz2026lagernvs}, $B{=}N$} & \refrow{26.43} & \refrow{0.793} & \refrow{0.271}\\
\bottomrule
\end{tabular}
\end{table}

\begin{table}[t]
\centering
\caption{Comparison with pose-free methods on Mip-NeRF 360~\cite{barron2022mipnerf360} with 9 source views (ReconFusion split~\cite{wu2024reconfusion}, 9 scenes), scored at $512\times512$ with the metric code of LagerNVS~\cite{szymanowicz2026lagernvs}.
Every method estimates the target pose from the target image and refines it on the rendered image.
Ours selects $B$ of the $N{=}12{,}321$ source tokens (1369 tokens are one view); with all tokens it reduces to LagerNVS~\cite{szymanowicz2026lagernvs} (gray).
Colors mark the \rankkey{} entry in each column, excluding the gray row.}
\label{tab:supp_full_mip360}
\footnotesize
\setlength{\tabcolsep}{5pt}
\renewcommand{\arraystretch}{1.1}
\begin{tabular}{@{}lccc@{}}
\toprule
Method & PSNR$\uparrow$ & SSIM$\uparrow$ & LPIPS$\downarrow$\\
\midrule
YoNoSplat~\cite{ye2026yonosplat} & 18.67 & 0.439 & 0.525\\ %
WorldMirror~\cite{liu2025worldmirror} & 17.71 & 0.418 & 0.502\\ %
\midrule
Ours, $B{=}1369$ & \trd 19.99 & \trd 0.484 & \trd 0.426\\
Ours, $B{=}2738$ & \snd 20.32 & \snd 0.494 & \snd 0.410\\
Ours, $B{=}4107$ & \fst 20.40 & \fst 0.496 & \fst 0.407\\ %
\midrule
\refrow{LagerNVS~\cite{szymanowicz2026lagernvs}, $B{=}N$} & \refrow{20.65} & \refrow{0.503} & \refrow{0.402}\\ %
\bottomrule
\end{tabular}
\end{table}

\section{ReSplat models}\label{sec:supp_resplat}
ReSplat~\cite{xu2026resplat} predicts an initial set of 3D Gaussians and then refines it over several iterations.
Its publicly available larger model has no refinement stage.
\Cref{tab:supp_resplat} compares the 8-view models on the 8-view ReSplat split~\cite{xu2026resplat}, with $256\times448$ inputs and the scoring of \cref{tab:supp_cmp_resplat}.
Without refinement, the larger model improves over the base model by 0.4\,dB.
With four refinement iterations, the base model reaches 29.14\,dB, 1.4\,dB above the larger model, and we therefore use it in the main text.
With square inputs, the ranking is the same, with 24.66, 25.02 and 27.78\,dB for the three rows of the table.

\begin{table}[t]
\centering
\caption{Publicly available 8-view models of ReSplat~\cite{xu2026resplat} on the 8-view ReSplat split of DL3DV~\cite{ling2024dl3dv} (140 scenes), with $256\times448$ inputs, scored as in \cref{tab:supp_cmp_resplat}.
The larger model is available only without refinement.}
\label{tab:supp_resplat}
\footnotesize
\setlength{\tabcolsep}{5pt}
\renewcommand{\arraystretch}{1.1}
\begin{tabular}{@{}llccc@{}}
\toprule
Model & Refinement & PSNR$\uparrow$ & SSIM$\uparrow$ & LPIPS$\downarrow$\\
\midrule
ReSplat-Base~\cite{xu2026resplat} & none & 27.29 & 0.882 & 0.115\\
ReSplat-Large~\cite{xu2026resplat} & none & 27.73 & 0.891 & 0.108\\
ReSplat-Base~\cite{xu2026resplat} & 4 iterations & \textbf{29.14} & \textbf{0.911} & \textbf{0.090}\\
\bottomrule
\end{tabular}
\end{table}

\section{Fine-tuning at the test view count}\label{sec:supp_finetune}
Our main model is trained with 2 to 6 source views.
On the ReSplat split~\cite{xu2026resplat}, whose models are each trained for a single view count, we fine-tune our router and renderer at each test view count for 2k iterations (\cref{sec:supp_details}).
\Cref{tab:supp_finetune} compares our model with and without this fine-tuning at 8 and 16 views.
Without fine-tuning, our model loses 0.04 to 0.5\,dB of PSNR.
It still outperforms ReSplat at 8 views with every budget, by 1.0 to 2.3\,dB, and at 16 views from two views' worth of tokens, as the fine-tuned model does.

\begin{table}[t]
\centering
\caption{Our model with and without fine-tuning at the test view count, on the ReSplat split~\cite{xu2026resplat} of DL3DV~\cite{ling2024dl3dv} (140 scenes).
Without fine-tuning, the model is the one trained with 2 to 6 source views.
Notation as in \cref{tab:supp_cmp_resplat}.}
\label{tab:supp_finetune}
\scriptsize
\setlength{\tabcolsep}{1.5pt}
\renewcommand{\arraystretch}{1.1}
\begin{tabular}{@{}lcccccc@{}}
\toprule
& \multicolumn{3}{c}{8 views} & \multicolumn{3}{c}{16 views}\\
\cmidrule(lr){2-4}\cmidrule(l){5-7}
Method & PSNR$\uparrow$ & SSIM$\uparrow$ & LPIPS$\downarrow$ & PSNR$\uparrow$ & SSIM$\uparrow$ & LPIPS$\downarrow$\\
\midrule
ReSplat$_{256\times448}$~\cite{xu2026resplat} & 29.14 & 0.911 & 0.090 & 28.98 & 0.908 & 0.092\\
\midrule
\multicolumn{7}{@{}l}{\emph{Without fine-tuning}}\\
Ours$_{256\times256}$, $B{=}1369$ & 30.16 & 0.922 & 0.060 & 28.18 & 0.888 & 0.078\\ %
Ours$_{256\times256}$, $B{=}2738$ & 31.32 & 0.934 & 0.049 & 29.47 & 0.906 & 0.061\\ %
Ours$_{256\times256}$, $B{=}4107$ & 31.48 & 0.936 & 0.047 & 29.67 & 0.908 & 0.059\\
\midrule
\multicolumn{7}{@{}l}{\emph{Fine-tuned at the test view count}}\\
Ours$_{256\times256}$, $B{=}1369$ & 30.27 & 0.923 & 0.058 & 28.68 & 0.896 & 0.072\\
Ours$_{256\times256}$, $B{=}2738$ & 31.37 & 0.935 & 0.048 & 29.86 & 0.911 & 0.058\\ %
Ours$_{256\times256}$, $B{=}4107$ & 31.51 & 0.936 & 0.047 & 30.00 & 0.913 & 0.057\\
\midrule
\refrow{LagerNVS$_{256\times256}$~\cite{szymanowicz2026lagernvs}, $B{=}N$} & \refrow{31.77} & \refrow{0.938} & \refrow{0.046} & \refrow{30.20} & \refrow{0.915} & \refrow{0.056}\\
\bottomrule
\end{tabular}
\end{table}

\section{Budgets below one view}\label{sec:supp_budgets}
The router can select any number of source tokens, including fewer than one view's worth.
\Cref{tab:supp_budgets_depthsplat,tab:supp_budgets_resplat} add budgets of 342 and 685 tokens, a quarter and a half of a view, to the comparisons of \cref{tab:supp_cmp_depthsplat,tab:supp_cmp_resplat}.
The renderer is adapted to budgets of at least 685 tokens, so 342 tokens lie outside the range seen in training.
Below one view's worth, quality drops quickly at every view count.
Halving the budget from one view's worth to 685 tokens costs 3.6 to 6.1\,dB of PSNR, and a quarter of a view costs 6.6 to 9.7\,dB.
With 685 tokens, our model is within 0.33\,dB of DepthSplat~\cite{xu2025depthsplat} but 5.0 to 5.8\,dB below ReSplat~\cite{xu2026resplat}.

\begin{table}[t]
\centering
\caption{Budgets below one view's worth of tokens on the DepthSplat split~\cite{xu2025depthsplat} of DL3DV~\cite{ling2024dl3dv} (140 scenes), with our main model, as in \cref{tab:supp_cmp_depthsplat}.
342 and 685 tokens are a quarter and a half of a view.
DepthSplat~\cite{xu2025depthsplat} receives $256\times448$ inputs.
Colors mark the \rankkey{} entry in each column, excluding the gray row.}
\label{tab:supp_budgets_depthsplat}
\scriptsize
\setlength{\tabcolsep}{1.5pt}
\renewcommand{\arraystretch}{1.1}
\begin{tabular}{@{}lcccccc@{}}
\toprule
& \multicolumn{3}{c}{4 views} & \multicolumn{3}{c}{6 views}\\
\cmidrule(lr){2-4}\cmidrule(l){5-7}
Method & PSNR$\uparrow$ & SSIM$\uparrow$ & LPIPS$\downarrow$ & PSNR$\uparrow$ & SSIM$\uparrow$ & LPIPS$\downarrow$\\
\midrule
DepthSplat$_{256\times448}$~\cite{xu2025depthsplat} & \trd 23.06 & \snd 0.786 & \snd 0.172 & \snd 23.87 & \snd 0.821 & \snd 0.147\\
\midrule
Ours$_{256\times256}$, $B{=}342$ & 20.23 & 0.651 & 0.284 & 20.20 & 0.652 & 0.287\\
Ours$_{256\times256}$, $B{=}685$ & \snd 23.18 & \trd 0.773 & \trd 0.175 & \trd 23.54 & \trd 0.787 & \trd 0.165\\
Ours$_{256\times256}$, $B{=}1369$ & \fst 26.79 & \fst 0.857 & \fst 0.105 & \fst 28.22 & \fst 0.889 & \fst 0.082\\ %
\midrule
\refrow{LagerNVS$_{256\times256}$~\cite{szymanowicz2026lagernvs}, $B{=}N$} & \refrow{27.60} & \refrow{0.871} & \refrow{0.094} & \refrow{29.49} & \refrow{0.906} & \refrow{0.068}\\
\bottomrule
\end{tabular}
\end{table}

\begin{table*}[t]
\centering
\caption{Budgets below one view's worth of tokens on the ReSplat split~\cite{xu2026resplat} of DL3DV~\cite{ling2024dl3dv} (140 scenes), with our model fine-tuned at each view count, as in \cref{tab:supp_cmp_resplat}.
342 and 685 tokens are a quarter and a half of a view.
ReSplat~\cite{xu2026resplat} receives $256\times448$ inputs.
Colors mark the \rankkey{} entry in each column, excluding the gray row.}
\label{tab:supp_budgets_resplat}
\scriptsize
\setlength{\tabcolsep}{3.2pt}
\renewcommand{\arraystretch}{1.1}
\begin{tabular}{@{}lccccccccc@{}}
\toprule
& \multicolumn{3}{c}{8 views} & \multicolumn{3}{c}{16 views} & \multicolumn{3}{c}{32 views}\\
\cmidrule(lr){2-4}\cmidrule(lr){5-7}\cmidrule(l){8-10}
Method & PSNR$\uparrow$ & SSIM$\uparrow$ & LPIPS$\downarrow$ & PSNR$\uparrow$ & SSIM$\uparrow$ & LPIPS$\downarrow$ & PSNR$\uparrow$ & SSIM$\uparrow$ & LPIPS$\downarrow$\\
\midrule
ReSplat$_{256\times448}$~\cite{xu2026resplat} & \snd 29.14 & \snd 0.911 & \snd 0.090 & \fst 28.98 & \fst 0.908 & \snd 0.092 & \fst 28.31 & \fst 0.900 & \snd 0.101\\ %
\midrule
Ours$_{256\times256}$, $B{=}342$ & 20.53 & 0.669 & 0.275 & 19.99 & 0.644 & 0.293 & 19.49 & 0.611 & 0.316\\ %
Ours$_{256\times256}$, $B{=}685$ & \trd 24.18 & \trd 0.810 & \trd 0.151 & \trd 23.34 & \trd 0.780 & \trd 0.167 & \trd 22.54 & \trd 0.741 & \trd 0.187\\
Ours$_{256\times256}$, $B{=}1369$ & \fst 30.27 & \fst 0.923 & \fst 0.058 & \snd 28.68 & \snd 0.896 & \fst 0.072 & \snd 26.91 & \snd 0.856 & \fst 0.090\\
\midrule
\refrow{LagerNVS$_{256\times256}$~\cite{szymanowicz2026lagernvs}, $B{=}N$} & \refrow{31.77} & \refrow{0.938} & \refrow{0.046} & \refrow{30.20} & \refrow{0.915} & \refrow{0.056} & \refrow{27.83} & \refrow{0.868} & \refrow{0.074}\\
\bottomrule
\end{tabular}
\end{table*}

\providecommand{\tbd}{\textcolor{red}{?}}
\providecommand{\refrow}[1]{\textcolor{gray}{#1}}
\section{Pose-free comparison: further methods and view counts}\label{sec:supp_posefree}
On RealEstate10K~\cite{zhou2018stereo} with 2 source views (\cref{tab:supp_posefree_re10k}), our model with one view's worth of tokens outperforms VicaSplat~\cite{li2025vicasplat} and FLARE~\cite{zhang2025flare}, each on the scenes it reports, by 2.2 and 2.4\,dB in PSNR.
With 3 and 6 source views on CO3D~\cite{reizenstein2021co3d} and Mip-NeRF 360~\cite{barron2022mipnerf360} (\cref{tab:supp_posefree_co3d,tab:supp_posefree_mip360}), it outperforms YoNoSplat~\cite{ye2026yonosplat} and WorldMirror~\cite{liu2025worldmirror} in all three metrics with every budget, by 1.4 to 3.6\,dB in PSNR, and with two views' worth of tokens it is 0.14 to 0.33\,dB below LagerNVS~\cite{szymanowicz2026lagernvs} with all tokens.

\begin{table}[t]
\centering
\caption{Further pose-free methods on RealEstate10K~\cite{zhou2018stereo} with 2 source views, under the protocol of \cref{tab:supp_full_re10k}.
VicaSplat~\cite{li2025vicasplat} reports on the whole NoPoSplat split~\cite{ye2025noposplat} (5,601 scenes); FLARE~\cite{zhang2025flare} reports on its subset (3,971 scenes) as the mean of the two overlap bins, and the rows below it are aggregated in the same way.
The numbers of VicaSplat~\cite{li2025vicasplat} and FLARE~\cite{zhang2025flare} are from their papers.
Colors mark the \rankkey{} entry in each column of each block, excluding the gray rows.}
\label{tab:supp_posefree_re10k}
\footnotesize
\setlength{\tabcolsep}{5pt}
\renewcommand{\arraystretch}{1.1}
\begin{tabular}{@{}lccc@{}}
\toprule
Method & PSNR$\uparrow$ & SSIM$\uparrow$ & LPIPS$\downarrow$\\
\midrule
\multicolumn{4}{@{}l}{\textit{NoPoSplat split~\cite{ye2025noposplat}}}\\
VicaSplat~\cite{li2025vicasplat} & \snd 25.04 & \snd 0.834 & \snd 0.161\\
Ours, $B{=}1369$ & \fst 27.27 & \fst 0.865 & \fst 0.142\\
\refrow{LagerNVS~\cite{szymanowicz2026lagernvs}, $B{=}N$} & \refrow{27.61} & \refrow{0.872} & \refrow{0.136}\\
\midrule
\multicolumn{4}{@{}l}{\textit{FLARE subset~\cite{zhang2025flare}, mean of the two overlap bins}}\\
FLARE~\cite{zhang2025flare} & \snd 23.77 & \trd 0.801 & \snd 0.191\\
YoNoSplat~\cite{ye2026yonosplat} & \trd 23.58 & \snd 0.803 & \trd 0.196\\ %
WorldMirror~\cite{liu2025worldmirror} & 21.75 & 0.773 & 0.230\\ %
Ours, $B{=}1369$ & \fst 26.13 & \fst 0.846 & \fst 0.160\\
\refrow{LagerNVS~\cite{szymanowicz2026lagernvs}, $B{=}N$} & \refrow{26.37} & \refrow{0.852} & \refrow{0.156}\\ %
\bottomrule
\end{tabular}
\end{table}

\begin{table}[t]
\centering
\caption{Comparison with pose-free methods on CO3D~\cite{reizenstein2021co3d} with 3 and 6 source views, under the protocol of \cref{tab:supp_full_co3d}.
$^\dagger$Trained on CO3D~\cite{reizenstein2021co3d}.
With 3 views, $N{=}4107$, so three views' worth of tokens is all of them and our model reduces to LagerNVS~\cite{szymanowicz2026lagernvs} (gray).
Colors mark the \rankkey{} entry in each column, excluding the gray row.}
\label{tab:supp_posefree_co3d}
\scriptsize
\setlength{\tabcolsep}{2.2pt}
\renewcommand{\arraystretch}{1.1}
\begin{tabular}{@{}lcccccc@{}}
\toprule
& \multicolumn{3}{c}{3 views} & \multicolumn{3}{c}{6 views}\\
\cmidrule(lr){2-4}\cmidrule(l){5-7}
Method & PSNR$\uparrow$ & SSIM$\uparrow$ & LPIPS$\downarrow$ & PSNR$\uparrow$ & SSIM$\uparrow$ & LPIPS$\downarrow$\\
\midrule
YoNoSplat~\cite{ye2026yonosplat} & \trd 19.18 & \trd 0.666 & 0.469 & 21.24 & 0.702 & 0.409\\
WorldMirror$^\dagger$~\cite{liu2025worldmirror} & 19.08 & 0.662 & \trd 0.458 & 20.66 & 0.702 & 0.404\\
\midrule
Ours, $B{=}1369$ & \snd 21.94 & \snd 0.715 & \snd 0.390 & \trd 24.29 & \trd 0.757 & \trd 0.321\\
Ours, $B{=}2738$ & \fst 22.17 & \fst 0.719 & \fst 0.384 & \snd 24.77 & \snd 0.765 & \snd 0.308\\
Ours, $B{=}4107$ & -- & -- & -- & \fst 24.83 & \fst 0.766 & \fst 0.306\\
\midrule
\refrow{LagerNVS~\cite{szymanowicz2026lagernvs}, $B{=}N$} & \refrow{22.31} & \refrow{0.722} & \refrow{0.379} & \refrow{25.10} & \refrow{0.772} & \refrow{0.300}\\
\bottomrule
\end{tabular}
\end{table}

\begin{table}[t]
\centering
\caption{Comparison with pose-free methods on Mip-NeRF 360~\cite{barron2022mipnerf360} with 3 and 6 source views, under the protocol of \cref{tab:supp_full_mip360}.
With 3 views, $N{=}4107$, so three views' worth of tokens is all of them and our model reduces to LagerNVS~\cite{szymanowicz2026lagernvs} (gray).
Colors mark the \rankkey{} entry in each column, excluding the gray row.}
\label{tab:supp_posefree_mip360}
\scriptsize
\setlength{\tabcolsep}{2.2pt}
\renewcommand{\arraystretch}{1.1}
\begin{tabular}{@{}lcccccc@{}}
\toprule
& \multicolumn{3}{c}{3 views} & \multicolumn{3}{c}{6 views}\\
\cmidrule(lr){2-4}\cmidrule(l){5-7}
Method & PSNR$\uparrow$ & SSIM$\uparrow$ & LPIPS$\downarrow$ & PSNR$\uparrow$ & SSIM$\uparrow$ & LPIPS$\downarrow$\\
\midrule
YoNoSplat~\cite{ye2026yonosplat} & \trd 16.83 & \trd 0.398 & 0.601 & 17.91 & 0.425 & 0.551\\ %
WorldMirror~\cite{liu2025worldmirror} & 16.35 & 0.378 & \trd 0.571 & 17.11 & 0.404 & 0.524\\ %
\midrule
Ours, $B{=}1369$ & \snd 18.25 & \snd 0.436 & \snd 0.510 & \trd 19.30 & \trd 0.469 & \trd 0.451\\ %
Ours, $B{=}2738$ & \fst 18.34 & \fst 0.438 & \fst 0.506 & \snd 19.49 & \snd 0.476 & \snd 0.441\\
Ours, $B{=}4107$ & -- & -- & -- & \fst 19.55 & \fst 0.477 & \fst 0.439\\
\midrule
\refrow{LagerNVS~\cite{szymanowicz2026lagernvs}, $B{=}N$} & \refrow{18.53} & \refrow{0.443} & \refrow{0.503} & \refrow{19.70} & \refrow{0.481} & \refrow{0.435}\\ %
\bottomrule
\end{tabular}
\end{table}

\section{Non-maximum suppression}\label{sec:supp_nms}
\begin{figure*}[t]
\centering
\includegraphics[width=\textwidth]{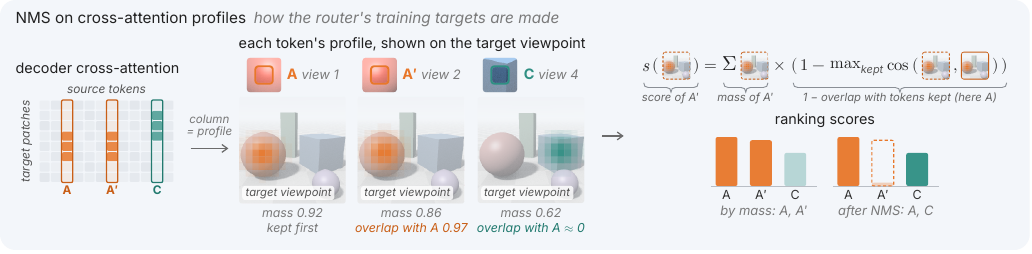}
\caption{\textbf{Non-maximum suppression (NMS) on tokens}, on a synthetic scene.
Ranking by attention mass keeps copies of the same content seen from several source views (A and A$'$), whereas NMS discounts the mass of each token by the overlap of its profile with those of the tokens already kept, so that the budget goes to new content (C).}
\label{fig:supp_nms_schematic}
\end{figure*}

\Cref{fig:supp_nms_schematic} illustrates the non-maximum suppression of \cref{sec:method_labels} on the synthetic scene of \cref{fig:method}, and \cref{fig:nms} shows its effect on a real scene.

\section{Label block}\label{sec:supp_labelblock}
Our router learns from the attention in the sixth of the twelve decoder blocks of LagerNVS~\cite{szymanowicz2026lagernvs} (\cref{sec:supp_details}), a choice we compare here with blocks 2, 4, 8 and 10.
Training a router for every block would be costly, so we instead select the tokens of each target viewpoint directly with the labels of each block, computed with the frozen decoder and all source tokens, and adapt the decoder to these selections for 20k iterations with the recipe of \cref{sec:supp_details}.
The five adaptations differ only in the block, with the same seed and the same sequence of scenes, source views, target viewpoints and budgets, and with block 6 the selections are exactly the labels that our router learns from.
We evaluate on 32 scenes of DL3DV-10K~\cite{ling2024dl3dv} that are excluded from all our training and from every benchmark split, though not from the training of LagerNVS, with 4, 8 and 16 source views spaced evenly over the first 300 frames and the same 60 target viewpoints per scene for every number of source views.

Block 6 gives the best adapted decoder with one view's worth of tokens in PSNR, SSIM and LPIPS (\cref{tab:supp_labelblock}).
It leads blocks 4 and 10 by 0.17 and 0.22\,dB, a lead that grows with the number of source views to 0.35 and 0.39\,dB with 16 views, and block 8 trails it by only 0.04\,dB, while the earliest block falls 1.3\,dB behind.
With two views' worth of tokens, the larger budget makes the choice of tokens less critical, so that blocks 4, 6 and 8 are on par, within 0.02\,dB of each other, and blocks 2 and 10 remain behind.

\begin{table}[t]
\centering
\caption{Decoders adapted to the selections made with the labels of different decoder blocks, on 32 held-out scenes of DL3DV-10K~\cite{ling2024dl3dv}, averaged over 4, 8 and 16 source views.
Our router learns from block 6.
Bold marks the best entry in each column, including ties at the shown precision.}
\label{tab:supp_labelblock}
\footnotesize
\setlength{\tabcolsep}{3.6pt}
\renewcommand{\arraystretch}{1.05}
\begin{tabular}{@{}lcccccc@{}}
\toprule
& \multicolumn{3}{c}{$B=1369$} & \multicolumn{3}{c}{$B=2738$}\\
\cmidrule(lr){2-4}\cmidrule(l){5-7}
Block & PSNR$\uparrow$ & SSIM$\uparrow$ & LPIPS$\downarrow$ & PSNR$\uparrow$ & SSIM$\uparrow$ & LPIPS$\downarrow$\\
\midrule
2 & 21.37 & 0.644 & 0.274 & 22.48 & 0.693 & 0.228\\
4 & 22.52 & 0.699 & 0.217 & \textbf{22.97} & \textbf{0.715} & 0.204\\
6 & \textbf{22.69} & \textbf{0.704} & \textbf{0.210} & 22.95 & 0.713 & \textbf{0.203}\\
8 & 22.65 & 0.703 & 0.211 & 22.94 & 0.714 & \textbf{0.203}\\
10 & 22.46 & 0.690 & 0.223 & 22.87 & 0.710 & 0.207\\
\bottomrule
\end{tabular}
\end{table}

\section{Selection strategies}\label{sec:supp_selection}
\Cref{tab:supp_selection} gives the numbers behind \cref{fig:selection}, repeats them with two views' worth of tokens, and adds two further strategies, the source view nearest to the target viewpoint and the selection rule of CLiFT~\cite{wang2025clift}.
Each strategy has its own decoder, adapted in the same way as ours (\cref{sec:supp_details}) with the selections of the strategy in place of those of the router, and no strategy sees the target image.
The strategies choose the source tokens as follows:
\begin{itemize}[leftmargin=*,topsep=2pt,itemsep=1pt]
\item \textbf{Ours without NMS} uses a router trained to rank the tokens by their attention mass alone (\cref{eq:mass}).
\item \textbf{Whole view} keeps the source view from which our router selects the most tokens.
\item \textbf{Nearest view} keeps the source view whose camera is nearest to the target viewpoint, measured with the distance in pose space of GP-MVS~\cite{hou2019gpmvs}, which combines translation and rotation; image-based renderers~\cite{wang2021ibrnet,suhail2022gpnr} likewise select their source views by proximity in pose, as do generative view synthesis models when they retrieve views from a memory~\cite{zhou2025seva}, each with its own distance.
\item \textbf{Geometric cost} renders the depth that MapAnything~\cite{keetha2025mapanything} estimates for the source views into the target image, computes for every visible source pixel the cost of Deep Blending~\cite{hedman2018deepblending}, a weighted sum of depth, viewing-angle and distance terms, and scores each token in each of the $37\times37$ cells of the target image by its largest cost over the pixels of the cell where it is visible.
It then fills the budget in rounds, each of which takes the best remaining token of every cell, as Inside-Out~\cite{hedman2016insideout} selects contributions per cell.
\item \textbf{CLiFT rule} applies the token selection that CLiFT~\cite{wang2025clift} performs at render time to our tokens: each patch of a coarse grid over the target image takes the source tokens closest to it, by the angle between their rays and the ray of the patch and by the distance between the cameras, and the rest of the budget is filled globally.
\item \textbf{Random tokens} are drawn uniformly from all source tokens for every target viewpoint.
\end{itemize}
With one view's worth of tokens, the whole view outperforms the nearest view by 0.1 to 0.2\,dB at every number of views, and the CLiFT rule~\cite{wang2025clift} trails the geometric cost by 0.2 to 0.8\,dB and outperforms random tokens by only 0.04\,dB with 4 views.
With two views' worth, the whole view and the nearest view keep a pair of views, our router again outperforms every other strategy up to 16 views, and the whole view is on par with it from 20 views on, within 0.05\,dB.

\begin{figure}[t]
\centering
\includegraphics[width=\columnwidth]{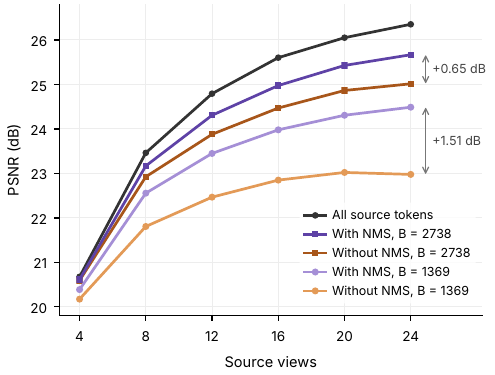}
\caption{\textbf{NMS as the source views become denser.}
PSNR of our model with routers trained with and without NMS, each with its own adapted renderer, at budgets of $B=1369$ and 2738 tokens, and of LagerNVS~\cite{szymanowicz2026lagernvs} with all source tokens, on the 140 test scenes of DL3DV~\cite{ling2024dl3dv} with 4 to 24 source views spaced evenly over the first 300 frames and the same 228 target viewpoints per scene.
Both routers are trained with 2 to 6 source views, and the arrows mark the gain of NMS at 24 views.}
\label{fig:supp_nms_views}
\end{figure}

\Cref{fig:supp_nms_views} compares our routers trained with and without NMS with two views' worth of tokens as well.
With two views' worth, the gain of NMS grows from 0.02 to 0.65\,dB between 4 and 24 views, and with 24 views, the router trained with NMS takes its tokens from 10.4 of the source views on average, against 5.9 for the router trained without it.

\begin{table*}[t]
\centering
\caption{Selection strategies with one and two views' worth of tokens ($B=1369$ and 2738) on the 140 test scenes of DL3DV~\cite{ling2024dl3dv}, with 4 to 24 source views (columns) spaced evenly over the first 300 frames and the same 228 target viewpoints per scene.
The left half gives the numbers behind \cref{fig:selection}.
Each strategy renders through a decoder adapted to its own selections, and with two views' worth, the whole view and the nearest view keep a pair of views.
Colors mark the \rankkey{} entry in each column of each block, excluding the gray rows.}
\label{tab:supp_selection}
\scriptsize
\setlength{\tabcolsep}{3.4pt}
\renewcommand{\arraystretch}{1.05}
\begin{tabular}{@{}lcccccccccccc@{}}
\toprule
& \multicolumn{6}{c}{One view's worth, $B=1369$} & \multicolumn{6}{c}{Two views' worth, $B=2738$}\\
\cmidrule(lr){2-7}\cmidrule(l){8-13}
Method & 4 & 8 & 12 & 16 & 20 & 24 & 4 & 8 & 12 & 16 & 20 & 24\\
\midrule
\multicolumn{13}{@{}l}{PSNR$\uparrow$}\\
Ours & \fst 20.39 & \fst 22.56 & \fst 23.45 & \fst 23.98 & \fst 24.31 & \fst 24.49 & \fst 20.61 & \fst 23.17 & \fst 24.31 & \fst 24.98 & \fst 25.43 & \snd 25.67\\
Ours without NMS & \snd 20.17 & \trd 21.81 & 22.47 & 22.85 & 23.02 & 22.98 & \snd 20.59 & \snd 22.92 & \trd 23.88 & 24.47 & 24.86 & 25.01\\
Whole view & 19.58 & 21.68 & \snd 22.76 & \snd 23.50 & \snd 24.05 & \snd 24.42 & 20.32 & \trd 22.80 & \snd 24.04 & \snd 24.81 & \snd 25.38 & \fst 25.71\\
Nearest view & 19.40 & 21.47 & 22.56 & \trd 23.36 & \trd 23.95 & \trd 24.33 & 20.28 & 22.62 & 23.79 & \trd 24.60 & \trd 25.20 & \trd 25.53\\
Geometric cost & \trd 20.08 & \snd 21.97 & \trd 22.59 & 22.81 & 22.87 & 22.87 & \trd 20.46 & 22.76 & 23.59 & 23.95 & 24.08 & 24.09\\
CLiFT rule~\cite{wang2025clift} & 19.32 & 21.34 & 22.11 & 22.44 & 22.62 & 22.69 & 20.09 & 22.19 & 23.04 & 23.45 & 23.69 & 23.79\\
Random tokens & 19.29 & 20.36 & 20.62 & 20.71 & 20.74 & 20.74 & 20.00 & 21.46 & 21.86 & 21.99 & 22.04 & 22.05\\
\refrow{LagerNVS~\cite{szymanowicz2026lagernvs}, $B{=}N$} & \refrow{20.68} & \refrow{23.46} & \refrow{24.79} & \refrow{25.60} & \refrow{26.05} & \refrow{26.35} & \refrow{20.68} & \refrow{23.46} & \refrow{24.79} & \refrow{25.60} & \refrow{26.05} & \refrow{26.35}\\
\midrule
\multicolumn{13}{@{}l}{SSIM$\uparrow$}\\
Ours & \fst 0.647 & \fst 0.732 & \fst 0.764 & \fst 0.781 & \fst 0.792 & \snd 0.798 & \fst 0.655 & \fst 0.749 & \fst 0.785 & \fst 0.805 & \fst 0.818 & \snd 0.825\\
Ours without NMS & \snd 0.640 & \trd 0.711 & 0.738 & 0.752 & 0.759 & 0.760 & \snd 0.654 & \snd 0.743 & \trd 0.776 & 0.795 & 0.807 & 0.812\\
Whole view & 0.626 & \snd 0.712 & \snd 0.750 & \snd 0.774 & \snd 0.790 & \fst 0.801 & 0.647 & \trd 0.740 & \snd 0.780 & \snd 0.802 & \snd 0.817 & \fst 0.826\\
Nearest view & 0.619 & 0.703 & \trd 0.742 & \trd 0.767 & \trd 0.785 & \trd 0.796 & 0.646 & 0.735 & 0.773 & \trd 0.796 & \trd 0.812 & \trd 0.821\\
Geometric cost & \trd 0.636 & 0.708 & 0.729 & 0.735 & 0.736 & 0.735 & \trd 0.651 & 0.737 & 0.765 & 0.775 & 0.779 & 0.779\\
CLiFT rule~\cite{wang2025clift} & 0.608 & 0.676 & 0.698 & 0.707 & 0.711 & 0.712 & 0.639 & 0.715 & 0.738 & 0.749 & 0.754 & 0.755\\
Random tokens & 0.588 & 0.622 & 0.629 & 0.630 & 0.629 & 0.628 & 0.625 & 0.672 & 0.682 & 0.684 & 0.685 & 0.684\\
\refrow{LagerNVS~\cite{szymanowicz2026lagernvs}, $B{=}N$} & \refrow{0.658} & \refrow{0.757} & \refrow{0.797} & \refrow{0.819} & \refrow{0.831} & \refrow{0.838} & \refrow{0.658} & \refrow{0.757} & \refrow{0.797} & \refrow{0.819} & \refrow{0.831} & \refrow{0.838}\\
\midrule
\multicolumn{13}{@{}l}{LPIPS$\downarrow$}\\
Ours & \fst 0.283 & \fst 0.198 & \fst 0.168 & \fst 0.153 & \fst 0.143 & \fst 0.137 & \fst 0.275 & \fst 0.181 & \fst 0.147 & \fst 0.130 & \fst 0.119 & \fst 0.111\\
Ours without NMS & \snd 0.292 & \trd 0.224 & 0.199 & 0.186 & 0.179 & 0.178 & \snd 0.276 & \snd 0.188 & \trd 0.158 & \trd 0.141 & 0.131 & 0.125\\ %
Whole view & 0.318 & 0.228 & \snd 0.190 & \snd 0.167 & \snd 0.151 & \snd 0.139 & 0.287 & 0.193 & \snd 0.156 & \snd 0.135 & \snd 0.122 & \snd 0.112\\
Nearest view & 0.332 & 0.241 & 0.201 & \trd 0.176 & \trd 0.158 & \trd 0.145 & 0.290 & 0.200 & 0.164 & 0.142 & \trd 0.127 & \trd 0.117\\
Geometric cost & \trd 0.293 & \snd 0.213 & \snd 0.190 & 0.183 & 0.180 & 0.179 & \trd 0.280 & \trd 0.190 & 0.161 & 0.150 & 0.145 & 0.143\\
CLiFT rule~\cite{wang2025clift} & 0.334 & 0.247 & 0.219 & 0.208 & 0.201 & 0.199 & 0.298 & 0.213 & 0.185 & 0.173 & 0.166 & 0.163\\
Random tokens & 0.340 & 0.293 & 0.284 & 0.281 & 0.280 & 0.279 & 0.304 & 0.246 & 0.233 & 0.229 & 0.227 & 0.226\\
\refrow{LagerNVS~\cite{szymanowicz2026lagernvs}, $B{=}N$} & \refrow{0.272} & \refrow{0.175} & \refrow{0.139} & \refrow{0.120} & \refrow{0.109} & \refrow{0.101} & \refrow{0.272} & \refrow{0.175} & \refrow{0.139} & \refrow{0.120} & \refrow{0.109} & \refrow{0.101}\\
\bottomrule
\end{tabular}
\end{table*}

\section{View switches}\label{sec:supp_view_switch}
\begin{figure*}[t]
\centering
\includegraphics[width=\textwidth]{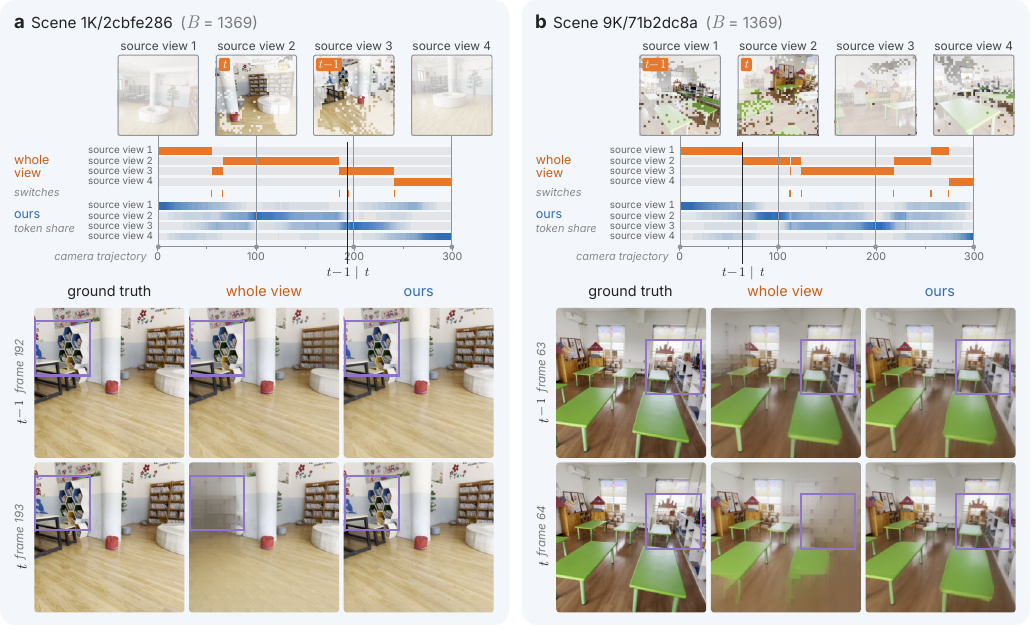}
\caption{Further view switches with one view's worth of tokens ($B{=}1369$), in the format of \cref{fig:view_switch}.
At the marked switch of the whole view, the content in the boxes vanishes (a) or turns translucent (b) in its rendering, whereas our router keeps 67\% (a) and 97\% (b) of its tokens and renders both frames alike.}
\label{fig:supp_view_switch_1369}
\end{figure*}

\begin{figure*}[t]
\centering
\includegraphics[width=\textwidth]{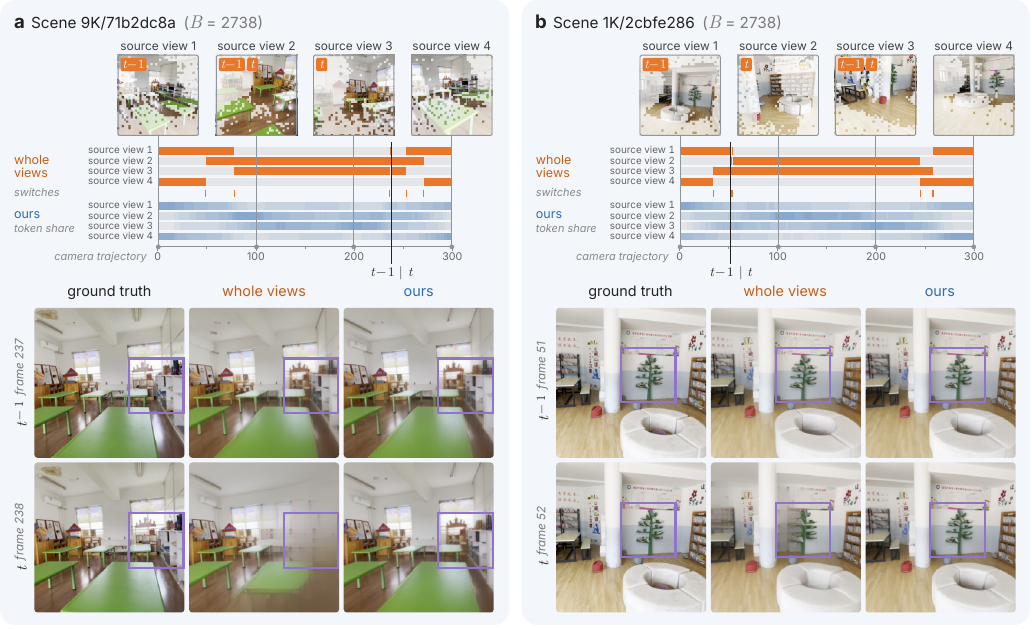}
\caption{Further view switches with two views' worth of tokens ($B{=}2738$), in the format of \cref{fig:view_switch}.
At the marked switch of the whole views, the content in the boxes turns translucent in their rendering, whereas our router keeps 97\% (a) and 99\% (b) of its tokens and renders both frames alike.}
\label{fig:supp_view_switch_2738}
\end{figure*}

\Cref{fig:supp_view_switch_1369,fig:supp_view_switch_2738} show further consecutive frames at which the whole view changes, in the format of \cref{fig:view_switch}.
In each example, content appears, vanishes or turns translucent in the rendering from the whole view at the switch, whereas our router keeps most of its tokens and renders both frames alike.

\section{Speed}\label{sec:supp_speed}
\Cref{tab:supp_speed} lists the measurements behind \cref{fig:speed}.
We time every setting on one otherwise idle RTX PRO 5000 GPU at $256\times256$, in bf16 and with one target viewpoint per frame, and capture the rendering of a frame once as a CUDA graph that is then replayed, for our model and for LagerNVS~\cite{szymanowicz2026lagernvs} alike.
For both models, all work that depends only on the scene is done once per scene and excluded from the time per frame: the encoding of the source views, the projections of the source tokens in the first decoder block and, for our model, the scene branch of the router.
The time per frame thus comprises the camera conditioning and the decoder, and for our model also the target branch of the router, the scoring of all source tokens and the selection of the top $B$; we report its median over 300 frames along the target viewpoints of each of three scenes.
The timed code renders the same images as the code used for evaluation, up to rounding differences that arise from processing one frame at a time.
At a fixed budget, the time per frame of our model still grows slightly with the number of source views, from 7.64 to 8.37\,ms between 4 and 24 views with 1369 tokens, because the router scores all source tokens for every frame.

\Cref{tab:supp_scene_cost} reports the work done once per scene.
The encoder, which LagerNVS~\cite{szymanowicz2026lagernvs} runs as well, takes 133\,ms with 4 source views and 1.37\,s with 24, and the scene branch of the router adds 29 to 44\% to this time.
The time that routing saves on every frame recovers the scene branch within 6 to 14 frames, and within 10 frames from 8 source views on.

We repeat the timing at $512\times512$ with the general model of LagerNVS~\cite{szymanowicz2026lagernvs}, which we use in the pose-free comparison, and with our router and adapted renderer for it (\cref{tab:supp_speed512}).
Since the time per frame does not depend on the content of the scene, we only time this model, with the scenes, source views, target poses and protocol used above; its source views also contribute 1369 tokens each.
With all tokens, the frame rate falls from 28.5\,FPS at 4 views to 7.7\,FPS at 24 views, and with 9 views, the largest number with which LagerNVS~\cite{szymanowicz2026lagernvs} reports real-time rendering on an H100 GPU, it reaches 17.9\,FPS on our GPU.
At a fixed budget, our model stays above 40\,FPS with 1369 tokens and above 33\,FPS with 2738 tokens at every number of views, and with 24 views it renders 5.3 and 4.3 times as many frames per second as LagerNVS~\cite{szymanowicz2026lagernvs}.
These speed-ups are smaller than at $256\times256$, because the four times larger number of target tokens raises the part of the cost per frame that does not depend on the source tokens.
With 24 views, budgets of 4107, 8214 and 16,428 tokens give 28.7, 22.3 and 14.0\,FPS, and the costs per scene are within 2\% of those in \cref{tab:supp_scene_cost}, since the encoder resizes its input views to the same size at both resolutions.

\Cref{tab:supp_speed_cmp} gives the speed of every method in \cref{tab:supp_cmp_depthsplat,tab:supp_cmp_resplat} and in \cref{tab:cmp_posefree}, timed on one otherwise idle RTX PRO 5000 GPU with one target viewpoint per call, at the resolution at which each table scores, and with the configuration that produced each method's numbers.
We time each Gaussian splatting method with its own renderer and with gsplat and the original rasterizer of 3D Gaussian Splatting, and the main text reports the fastest renderer that reproduces the method's images, with a PSNR above 50\,dB or a largest difference below 1/255 against its own renderer.
CLiFT~\cite{wang2025clift} renders in half precision from a CUDA graph, as our model and LagerNVS~\cite{szymanowicz2026lagernvs} do, and renders its full $256\times448$ view, from which the scored square is taken.
The time per scene covers the prediction of the Gaussians, including the four refinement iterations of ReSplat~\cite{xu2026resplat}, the encoder and the clustering of CLiFT, and the encoder of LagerNVS together with, for our model, the scene branch of the router.
The pose refinement that precedes rendering in the pose-free comparison is not included and takes 6.1 to 10.1\,s per target viewpoint for our model, 7.3 to 18.8\,s for LagerNVS and 0.8 to 7.5\,s for the Gaussian splatting methods.

\begin{table}[t]
\centering
\caption{Speed and quality behind \cref{fig:speed}, on the 140 test scenes of DL3DV~\cite{ling2024dl3dv} with the source views spaced evenly over the first 300 frames and 228 target viewpoints per scene.
$B$ is the number of source tokens per frame (1369 per view), and the gray rows use all $N$ source tokens, with which our model reduces to LagerNVS~\cite{szymanowicz2026lagernvs}.
The time per frame (ms) is the median over 900 frames on an RTX PRO 5000 GPU, and FPS is its inverse.
$^\dagger$Larger than any budget used in training.}
\label{tab:supp_speed}
\footnotesize
\setlength{\tabcolsep}{4.5pt}
\renewcommand{\arraystretch}{1.05}
\begin{tabular}{@{}rrrrccc@{}}
\toprule
Views & $B$ & ms & FPS & PSNR$\uparrow$ & SSIM$\uparrow$ & LPIPS$\downarrow$\\
\midrule
4 & 1369 & 7.64 & 130.9 & 20.39 & 0.647 & 0.283\\
 & 2738 & 11.48 & 87.1 & 20.61 & 0.655 & 0.275\\
 & \refrow{$N$} & \refrow{14.35} & \refrow{69.7} & \refrow{20.68} & \refrow{0.658} & \refrow{0.272}\\
\midrule
8 & 1369 & 7.80 & 128.2 & 22.56 & 0.732 & 0.198\\
 & 2738 & 11.65 & 85.8 & 23.17 & 0.749 & 0.181\\
 & \refrow{$N$} & \refrow{24.71} & \refrow{40.5} & \refrow{23.46} & \refrow{0.757} & \refrow{0.175}\\
\midrule
12 & 1369 & 7.97 & 125.5 & 23.45 & 0.764 & 0.168\\
 & 2738 & 11.83 & 84.5 & 24.31 & 0.785 & 0.147\\
 & \refrow{$N$} & \refrow{35.70} & \refrow{28.0} & \refrow{24.79} & \refrow{0.797} & \refrow{0.139}\\
\midrule
16 & 1369 & 8.10 & 123.5 & 23.98 & 0.781 & 0.153\\
 & 2738 & 11.97 & 83.6 & 24.98 & 0.805 & 0.130\\
 & \refrow{$N$} & \refrow{49.26} & \refrow{20.3} & \refrow{25.60} & \refrow{0.819} & \refrow{0.120}\\
\midrule
20 & 1369 & 8.18 & 122.2 & 24.31 & 0.792 & 0.143\\
 & 2738 & 12.07 & 82.9 & 25.43 & 0.818 & 0.119\\
 & \refrow{$N$} & \refrow{63.94} & \refrow{15.6} & \refrow{26.05} & \refrow{0.831} & \refrow{0.109}\\
\midrule
24 & 1369 & 8.37 & 119.4 & 24.49 & 0.798 & 0.137\\
 & 2738 & 12.27 & 81.5 & 25.67 & 0.825 & 0.111\\
 & 4107 & 15.28 & 65.4 & 25.93 & 0.830 & 0.106\\
 & 8214$^\dagger$ & 20.38 & 49.1 & 26.07 & 0.833 & 0.104\\
 & 16,428$^\dagger$ & 37.27 & 26.8 & 26.22 & 0.835 & 0.103\\
 & \refrow{$N$} & \refrow{78.76} & \refrow{12.7} & \refrow{26.35} & \refrow{0.838} & \refrow{0.101}\\
\bottomrule
\end{tabular}
\end{table}

\begin{table}[t]
\centering
\caption{Work done once per scene, in ms on an RTX PRO 5000 GPU, and the number of frames after which the time that routing saves per frame exceeds the time of the scene branch of the router.
The encoder is shared by our model and LagerNVS~\cite{szymanowicz2026lagernvs}.}
\label{tab:supp_scene_cost}
\footnotesize
\setlength{\tabcolsep}{5pt}
\renewcommand{\arraystretch}{1.05}
\begin{tabular}{@{}rrrcc@{}}
\toprule
& & & \multicolumn{2}{c}{Frames to recover}\\
\cmidrule(l){4-5}
Views & Encoder & Router & $B=1369$ & $B=2738$\\
\midrule
4 & 133 & 39 & 6 & 14\\
8 & 303 & 101 & 6 & 8\\
12 & 520 & 189 & 7 & 8\\
16 & 784 & 299 & 8 & 9\\
20 & 1058 & 433 & 8 & 9\\
24 & 1371 & 600 & 9 & 10\\
\bottomrule
\end{tabular}
\end{table}

\begin{table}[t]
\centering
\caption{Time per frame (ms) and frames per second at $512\times512$, for the general model of LagerNVS~\cite{szymanowicz2026lagernvs} with all tokens and for our model built on it, on an RTX PRO 5000 GPU with the protocol of \cref{tab:supp_speed}.
Our model is not timed with 9 views.}
\label{tab:supp_speed512}
\footnotesize
\setlength{\tabcolsep}{3.6pt}
\renewcommand{\arraystretch}{1.05}
\begin{tabular}{@{}rcccccc@{}}
\toprule
& \multicolumn{2}{c}{All tokens} & \multicolumn{2}{c}{Ours, $B=1369$} & \multicolumn{2}{c}{Ours, $B=2738$}\\
\cmidrule(lr){2-3}\cmidrule(lr){4-5}\cmidrule(l){6-7}
Views & ms & FPS & ms & FPS & ms & FPS\\
\midrule
4 & 35.08 & 28.5 & 23.61 & 42.4 & 29.15 & 34.3\\
8 & 51.88 & 19.3 & 23.88 & 41.9 & 29.41 & 34.0\\
9 & 55.90 & 17.9 & -- & -- & -- & --\\
12 & 69.64 & 14.4 & 24.11 & 41.5 & 29.63 & 33.8\\
16 & 88.72 & 11.3 & 24.25 & 41.2 & 29.75 & 33.6\\
20 & 108.86 & 9.2 & 24.37 & 41.0 & 29.86 & 33.5\\
24 & 129.44 & 7.7 & 24.56 & 40.7 & 30.06 & 33.3\\
\bottomrule
\end{tabular}
\end{table}

\begin{table}[t]
\centering
\caption{Time per frame and per scene for the methods of \cref{tab:supp_cmp_depthsplat,tab:supp_cmp_resplat} and of \cref{tab:cmp_posefree}, on one RTX PRO 5000 GPU with one target viewpoint per call.
For the Gaussian splatting methods, the time per frame is given with the fastest renderer that reproduces their images and, in parentheses, with their own renderer; for CLiFT~\cite{wang2025clift}, the parentheses give its own code without a CUDA graph.
On RealEstate10K, our model and LagerNVS~\cite{szymanowicz2026lagernvs} render at $512\times512$ and downsample, as in the evaluation.}
\label{tab:supp_speed_cmp}
\footnotesize
\setlength{\tabcolsep}{3.6pt}
\renewcommand{\arraystretch}{1.05}
\begin{tabular}{@{}llccr@{}}
\toprule
Method & Views & Frame (ms) & FPS & Scene (ms)\\
\midrule
\multicolumn{5}{@{}l}{\emph{DL3DV, DepthSplat split, $256\times256$}}\\
DepthSplat~\cite{xu2025depthsplat} & 4 & 1.86 (2.29) & 538 & 77\\
CLiFT~\cite{wang2025clift} & 4 & 7.03 (12.07) & 142 & 272\\
Ours, $B{=}1369$ & 4 & 7.73 & 129 & 172\\
Ours, $B{=}2738$ & 4 & 11.49 & 87 & 172\\
Ours, $B{=}4107$ & 4 & 14.58 & 69 & 172\\ %
LagerNVS~\cite{szymanowicz2026lagernvs} & 4 & 14.49 & 69 & 134\\
DepthSplat~\cite{xu2025depthsplat} & 6 & 2.27 (2.79) & 440 & 106\\
CLiFT~\cite{wang2025clift} & 6 & 8.57 (15.71) & 117 & 521\\
Ours, $B{=}1369$ & 6 & 7.81 & 128 & 278\\
Ours, $B{=}2738$ & 6 & 11.61 & 86 & 278\\
Ours, $B{=}4107$ & 6 & 14.54 & 69 & 278\\
LagerNVS~\cite{szymanowicz2026lagernvs} & 6 & 19.08 & 52 & 213\\
\midrule
\multicolumn{5}{@{}l}{\emph{DL3DV, ReSplat split, $256\times256$}}\\
ReSplat~\cite{xu2026resplat} & 8 & 0.59 (0.89) & 1684 & 260\\
Ours, $B{=}1369$ & 8 & 7.87 & 127 & 405\\
Ours, $B{=}2738$ & 8 & 11.69 & 86 & 405\\
Ours, $B{=}4107$ & 8 & 14.76 & 68 & 405\\
LagerNVS~\cite{szymanowicz2026lagernvs} & 8 & 24.83 & 40 & 304\\
ReSplat~\cite{xu2026resplat} & 16 & 0.70 (0.99) & 1436 & 566\\
Ours, $B{=}1369$ & 16 & 8.20 & 122 & 1084\\
Ours, $B{=}2738$ & 16 & 12.08 & 83 & 1084\\
Ours, $B{=}4107$ & 16 & 15.10 & 66 & 1084\\
LagerNVS~\cite{szymanowicz2026lagernvs} & 16 & 49.74 & 20 & 791\\
ReSplat~\cite{xu2026resplat} & 32 & 1.00 (1.30) & 996 & 1302\\
Ours, $B{=}1369$ & 32 & 8.76 & 114 & 3106\\
Ours, $B{=}2738$ & 32 & 12.65 & 79 & 3106\\
Ours, $B{=}4107$ & 32 & 15.67 & 64 & 3106\\
LagerNVS~\cite{szymanowicz2026lagernvs} & 32 & 106.70 & 9 & 2114\\
\midrule
\multicolumn{5}{@{}l}{\emph{RealEstate10K, $256\times256$}}\\
NoPoSplat~\cite{ye2025noposplat} & 2 & 1.15 (1.71) & 868 & 33\\
SPFSplatV2~\cite{huang2026spfsplatv2} & 2 & 1.21 (1.88) & 828 & 36\\
YoNoSplat~\cite{ye2026yonosplat} & 2 & 1.15 (2.08) & 867 & 57\\
WorldMirror~\cite{liu2025worldmirror} & 2 & 1.18 (1.44) & 847 & 106\\
Ours, $B{=}1369$ & 2 & 24.12 & 41 & 94\\
LagerNVS~\cite{szymanowicz2026lagernvs} & 2 & 29.69 & 34 & 78\\
\midrule
\multicolumn{5}{@{}l}{\emph{CO3D, $512\times512$}}\\
YoNoSplat~\cite{ye2026yonosplat} & 9 & 2.03 (5.72) & 493 & 92\\
WorldMirror~\cite{liu2025worldmirror} & 9 & 1.90 (2.24) & 527 & 465\\
Ours, $B{=}1369$ & 9 & 24.05 & 42 & 472\\
Ours, $B{=}2738$ & 9 & 29.90 & 33 & 472\\
Ours, $B{=}4107$ & 9 & 34.46 & 29 & 472\\
LagerNVS~\cite{szymanowicz2026lagernvs} & 9 & 57.19 & 17 & 354\\
\midrule
\multicolumn{5}{@{}l}{\emph{Mip-NeRF 360, $512\times512$}}\\
YoNoSplat~\cite{ye2026yonosplat} & 9 & 1.98 (4.80) & 504 & 92\\
WorldMirror~\cite{liu2025worldmirror} & 9 & 2.01 (2.33) & 498 & 467\\
Ours, $B{=}1369$ & 9 & 24.01 & 42 & 468\\
Ours, $B{=}2738$ & 9 & 29.92 & 33 & 468\\
Ours, $B{=}4107$ & 9 & 34.22 & 29 & 468\\
LagerNVS~\cite{szymanowicz2026lagernvs} & 9 & 56.84 & 18 & 352\\
\bottomrule
\end{tabular}
\end{table}

\section{Refining the target pose with input poses}\label{sec:supp_posed_refine}
On the ReSplat split (\cref{tab:supp_cmp_resplat}), our model is ahead of ReSplat~\cite{xu2026resplat} in PSNR and SSIM with 8 views but behind in four of the six settings with 16 and 32 views, whereas its LPIPS is lower in all nine settings, by 0.011 to 0.043.
We attribute this to a small misalignment of our renders.
Like LagerNVS~\cite{szymanowicz2026lagernvs}, our model takes the target viewpoint as an input and renders it in a coordinate system that the network constructs implicitly, rather than by rasterizing explicit Gaussians with the given camera, as ReSplat does.
The rendered image can therefore be slightly shifted or warped relative to the target image.
Such shifts are hard to see but are penalized heavily by pixel-wise metrics such as PSNR and SSIM, and much less by LPIPS.
LagerNVS shows the same pattern: with 32 views, its PSNR is 0.48\,dB below that of ReSplat, while its LPIPS is 0.074 against 0.101.

To test this, we refine the target pose on the target image as in the pose-free protocol (\cref{sec:supp_details}), starting from the ground-truth pose (\cref{tab:supp_posed_refine}).
Our router selects the tokens once from the ground-truth pose and the selection is held fixed, and ReSplat keeps its Gaussians and optimizes only the camera.

Refinement raises the PSNR of our model by 0.6 to 1.9\,dB, more with more source views, whereas that of ReSplat changes by $+0.06$, $-0.02$ and $-0.17$\,dB with 8, 16 and 32 views.
With refinement, our model is ahead of ReSplat in PSNR and LPIPS in all nine settings, in PSNR by 0.27 to 3.01\,dB, and in SSIM in all but one ($B{=}1369$ with 32 views).
The corrections are small: for our model, the median rotation is 0.04 to 0.10 degrees and the median translation is 0.6 to 2.0\% of the median distance between consecutive source cameras.
That ReSplat gains at most 0.06\,dB indicates that the target poses are consistent with the source poses, so the correction compensates for the misalignment of our renders.

\begin{table*}[t]
\centering
\caption{The comparison with input poses of \cref{tab:supp_cmp_resplat}, on the ReSplat split~\cite{xu2026resplat} of DL3DV~\cite{ling2024dl3dv} (140 scenes), repeated with the target pose refined on the target image as in the pose-free protocol, starting from the ground-truth pose (\cref{sec:supp_posed_refine}).
ReSplat~\cite{xu2026resplat} receives $256\times448$ inputs, and all methods are scored on the central $256\times256$ square, as in \cref{tab:supp_cmp_resplat}, which gives the numbers without refinement.
LagerNVS~\cite{szymanowicz2026lagernvs} is not refined in this experiment.
The three view counts use different target viewpoints, so numbers are comparable only within a view count.
Colors mark the \rankkey{} entry in each column.}
\label{tab:supp_posed_refine}
\footnotesize
\setlength{\tabcolsep}{6pt}
\renewcommand{\arraystretch}{1.1}
\begin{tabular}{@{}lccccccccc@{}}
\toprule
& \multicolumn{3}{c}{8 views} & \multicolumn{3}{c}{16 views} & \multicolumn{3}{c}{32 views}\\
\cmidrule(lr){2-4}\cmidrule(lr){5-7}\cmidrule(l){8-10}
Method & PSNR$\uparrow$ & SSIM$\uparrow$ & LPIPS$\downarrow$ & PSNR$\uparrow$ & SSIM$\uparrow$ & LPIPS$\downarrow$ & PSNR$\uparrow$ & SSIM$\uparrow$ & LPIPS$\downarrow$\\
\midrule
ReSplat$_{256\times448}$~\cite{xu2026resplat} & 29.20 & 0.911 & 0.090 & 28.96 & 0.907 & 0.092 & 28.15 & \trd 0.895 & 0.102\\
\midrule
Ours$_{256\times256}$, $B{=}1369$ & \trd 30.89 & \trd 0.930 & \trd 0.056 & \trd 29.67 & \trd 0.913 & \trd 0.067 & \trd 28.42 & 0.889 & \trd 0.080\\
Ours$_{256\times256}$, $B{=}2738$ & \snd 32.08 & \snd 0.941 & \snd 0.046 & \snd 31.06 & \fst 0.928 & \snd 0.053 & \snd 29.91 & \snd 0.910 & \snd 0.063\\
Ours$_{256\times256}$, $B{=}4107$ & \fst 32.21 & \fst 0.942 & \fst 0.045 & \fst 31.19 & \fst 0.928 & \fst 0.052 & \fst 30.10 & \fst 0.912 & \fst 0.061\\
\bottomrule
\end{tabular}
\end{table*}

\end{document}